\documentclass[review]{elsarticle}

\usepackage{amssymb, amsmath, amsthm, amsfonts}
\usepackage{booktabs} % For prettier tables
\usepackage{arydshln}
\usepackage{graphicx}
\usepackage[config, position=bottom]{subfig}
\usepackage{url}
\usepackage{tikz}
\usepackage{multirow}
\usepackage[pdftex,
            pdfauthor={Dickson Owuor, Thomas Runkler, Anne Laurent},
            pdftitle={Swarm Based Optimization for Mining Gradual Patterns},
            pdfsubject={Data Mining, Evolutionary Algorithms},
            pdfkeywords={Generic Algorithm, Particle Swarm Optimization, Wasp Swarm Optimization, Random Search, Local Search}]{hyperref}
\usetikzlibrary{positioning,shapes.misc}

\usepackage{lineno,hyperref}
\modulolinenumbers[5]

\journal{Journal of Swarm and Evolutionary Computation}

\begin{document}

\begin{frontmatter} % For elsarticle

\title{A Metaheuristic Approach for Mining Gradual Patterns}

\author[addr1]{Dickson Odhiambo Owuor\corref{corrauthor}}
\cortext[corrauthor]{Corresponding author}
\ead{dowuor@strathmore.edu}

\author[addr2]{Thomas Runkler}
\ead{thomas.runkler@siemens.com}

\author[addr3]{Anne Laurent}
\ead{anne.laurent@umontpellier.fr}

\address[addr1]{SCES Strathmore University, Nairobi, Kenya}
\address[addr2]{Siemens AG, Munich, Germany}
\address[addr3]{LIRMM Univ Montpellier, CNRS, Montpellier, France}

\begin{abstract}
Swarm intelligence is a discipline that studies the collective behavior that is produced by local interactions of a group of individuals with each other and with their environment. In Computer Science domain, numerous swarm intelligence techniques are applied to optimization problems that seek to efficiently find best solutions within a search space. Gradual pattern mining is another Computer Science field that could benefit from the efficiency of swarm based optimization techniques in the task of finding gradual patterns from a huge search space. A gradual pattern is a rule-based correlation that describes the gradual relationship among the attributes of a data set. For example, given attributes $\{G, H\}$ of a data set a gradual pattern may take the form: \textit{``the less G, the more H''} (simply denoted as $\{G^{\downarrow},H^{\uparrow} \}$). In this paper, we propose a numeric encoding for gradual pattern candidates that we use to define an effective search space. In addition, we present a systematic study of several meta-heuristic optimization techniques as efficient solutions to the problem of finding gradual patterns using our search space.
\end{abstract}

 
\end{frontmatter} % For elsarticle

	%\linenumbers
	\section{Introduction}
	\label{sec:introduction}
	Swarm-based optimization algorithms, most of which mimic the collective intelligence of groups of simple agents, have demonstrated excellent efficiency in solving many optimization problems \cite{Wahab2015}. According to \cite{Mohamed2018}, swarm-based optimization algorithms belong to the subcategory of metaphor-based algorithms in the class of metaheuristic algorithms.
	
	Metaheuristic algorithms are computational intelligence models that are used to solve complex optimization problems. In this paper, we focus on the task of generating gradual pattern candidates as an optimization problem. Gradual pattern mining is a data mining field in Computer Science that deals with describing attributes/variables/features of data sets using rules such as: \textit{``the higher the temperature, the less vehicle traffic'' }. We provide a detailed description of gradual pattern mining in Section~\ref{sec:background}.
	
	Gradual pattern (GP) mining has many applications especially in domains that seek to find correlational knowledge about variables of a data set. For instance, in \cite{Chen2018}, GP mining is applied in the discovery of the gradual relationship between numerical building variables in order to improve building efficiency.
	
	One key challenge experienced in GP mining is in the task of identifying candidate rules. This involves combining different kinds of variations (increasing and decreasing) for each rule that correlates two or more variables \cite{Di-Jorio2008,Owuor2021jul}. However, the task of identifying these candidates may be modelled into a search space problem onto which metaheuristic algorithms may be applied.
	
	In this paper, we propose a search space such that each point is a numeric digit that encodes a GP candidate. Our experiment results show that our numeric encoding produces a lean search space metaheuristic algorithms to explore. The results also rank the computational efficiency of these algorithms in the task of finding \textit{best} GP candidates from the search space.
	
	The main contributions of this study are described as follows:
	\begin{itemize}
		\item We propose a numeric encoding that allows us to build a lean search space for GP candidates.
		\item We develop metaheuristic algorithms and compare their performances in exploring search spaces based on our numeric encoding and those based on other encodings.
	\end{itemize}

	\section{Background}
	\label{sec:background}
	In Computer Science, a search space may be described as \textit{``an expanse that is defined by the set of all feasible solutions through which an algorithm may iterate''} \cite{Obitko1999,Patel2021}. In GP mining, a search space (known to be a lattice) holds all possible candidates that may be validated to be relevant GPs.
	
	The subject of mining GPs (also referred to as gradual rules) has been studied by many researchers. To begin with, there exists different notions of gradual rules. Of particular interest, is the notion of applying Rescher-Gaines implication in order to measure gradualness. For instance, given two attributes $A_{1}$ and $A_{2}$, fuzzy sets $M$ and $N$ defined on $A_{1}$ and $A_{2}$ respectively. If $M(A_{1})$ be the membership degree of $A_{1}$ in $M$, then the rule can be expresses as \textit{``$A_{1}$ is in $M$ implies $A_{2}$ is in $N$''}. In such expressions, we use Rescher-Gaines implication:
		
	\begin{equation}\label{eqn:rescher-gaines}
		A_{1} \rightarrow_{RG} A_{2} = 
		\begin{cases}
			1 \quad $if $ M(A_{1}) \leq N(A_{2})\\
			0 \quad $else$\\
		\end{cases}
	\end{equation}
	
	Starting from the above notions, \cite{Berzal2007} formalizes 2 main kinds of gradual rules of the form  \textit{``the more/less $A_{1}$ is in $M$, the more/less $A_{2}$ is in $N$''.} The first kind (known as pure gradual rules), verifies a rule by evaluating properties between individual objects. The second kind (known as gradual dependence) verifies a rule by comparing the properties between objects. In this paper, we deal with pure gradual rules.
	
	For the purpose of facilitating comprehension of this paper, we use a sample data set with 3 attributes and 4 objects or tuples (as shown in Table~\ref{tab:sample_1a}) to introduce GP mining. For instance, using these 3 attributes  $\{age, sessions, marks \}$ we may formulate approximately 20 GP candidates like $\{age^{\uparrow}, sessions^{\downarrow}\}$,\\$\{sessions^{\downarrow}, marks^{\uparrow}\}$, $\{age^{\uparrow}, sessions^{\downarrow}, marks^{\uparrow}\}$ among others.
	
	\begin{table}[!h]
	\centering
	\footnotesize
	\caption{A sample data set showing details of students who took part in a graded course.}
	\label{tab:sample_1a}
	\begin{tabular}{c c c}
	\textbf{age} & \textbf{sessions} & \textbf{marks}\\
	\hline \hline
	23 & 2 & 55\\
	32 & 4 & 64\\
	40 & 5 & 78\\
	25 & 5 & 48\\
	\bottomrule
	\end{tabular}
	\end{table}
	
	To clarify, only GP candidates whose computed \textit{frequency support} exceed a specified \textit{user-defined threshold} may be verified as relevant GPs \cite{Laurent2009,Owuor2021aug}. Generally, frequency support may be defined as: \textit{``the proportion of ordered objects in a data set that verify a particular GP''}. For instance, the GP candidate $\{sessions^{\uparrow}, marks^{\uparrow}\}$ is verified by the first 3 objects (out of the 4 objects) of the data set (shown in Table~\ref{tab:sample_1a}) if they are taken in ascending order.
	
	This is due to the fact that the object values of the attributes $\{sessions$, $marks\}$ subsequently increase in the first 3 objects with the exception of the fourth object values. Therefore, in this case the frequency support may be computed to be 3/4 or 0.75.
	
	In view of this example, it is noticeable that the number of GP candidates is directly proportional to the number of a data set's attributes. According to \cite{Di-Jorio2008,Owuor2021aug} (in GP mining), deterministic algorithms that have to find and validate each GP candidate are computationally overwhelmed when applied on data sets with large attribute sizes.
	
	However, an effective search space $\mathcal{S}_{gp}$ may be defined such that every point in the search space represents a possible GP candidate. In addition, it is possible to iterate through the search space using a non-deterministic algorithm to find candidates which are validated by computing their frequency support. %The candidate with the highest frequency support becomes the \textit{best candidate}. %Such non-deterministic algorithms may be more computationally efficient than their deterministic counterparts at finding best GPs, since they do not have to validate every possible GP candidate.

	\subsubsection*{Definitions and Notations}
	
	We give some definitions of gradual patterns as provided in \cite{Owuor2021jul,Laurent2009,Owuor2019,Owuor2021oct,Di-Jorio2009}. For example, let D be a data set with attributes \textit{\{age, sessions, marks\}} and objects \textit{\{obj1, obj2, obj3, obj4\}} (as shown in Table~\ref{tab:sample_1b}).
	
	\begin{table}[!h]
	\centering
	\footnotesize
	\caption{A sample data set, D, showing details of participants who took part in a training.}
	\label{tab:sample_1b}
	\begin{tabular}{c|c c c}
	\textbf{obj\#} & \textbf{age} & \textbf{sessions} & \textbf{marks}\\
	\hline \hline
	obj1 & 23 & 2 & 55\\
	obj2 & 32 & 4 & 64\\
	obj3 & 40 & 5 & 78\\
	obj4 & 25 & 5 & 48\\
	\bottomrule
	\end{tabular}
	\end{table}

	\textbf{Definition 1.1.} \texttt{(Gradual Item).} \textit{A gradual item is a pair $at^{v}$: where $at$ is an attribute and $v$ is a gradual variation such that $v \in \{\uparrow, \downarrow\}$. $\uparrow$ denotes an increasing variation and $\downarrow$ denotes a decreasing variation.}
	
	For example, $age^{\uparrow}$ can be interpreted as \textit{``the higher the age''.}
			
	\textbf{Definition 1.2.} \texttt{(Gradual Pattern).} \textit{A gradual pattern GP is a set of gradual items such that $GP = \{at_{1}^{v}, ..., at_{n}^{v}\}$.}
	
	For example, $\{age^{\uparrow}, sessions^{\downarrow}, marks^{\uparrow}\}$ is a GP that can be interpreted as \textit{``the higher the age, the less session numbers, the higher the marks obtained.''}
	
	The quality of a gradual item or gradual pattern is measured by its \textit{frequency support.} The frequency support of a gradual pattern $sup(GP)$ may be defined as: \textit{``the proportion of object couples that respect the gradual variations given collectively by all the items in the pattern''.} For instance, let $A$ be an attribute of data set $\mathcal{D}$ which has a total of $k$ objects and $o$ be an object in $\mathcal{D}$ such that $A(o)$ denotes the value $A$ takes for $o$, $(o,o')$ denotes an object pair in $\mathcal{D}'$ (which is a transaction data set derived from $\mathcal{D}$ and it holds the set of all object pair combinations). The gradual item $A^{\downarrow}$ holds if $\forall (o,o') \in \mathcal{D}',~ A(o) > A(o')$ \cite{Laurent2009,Clementin2021}. Therefore, support of a GP can be defined by the formula that follows.
	
	%\begin{equation}\label{eqn:support}
	%	sup(GP) = \frac{1}{|\mathcal{D}'|}\cdot |\{\forall(o,o')\in \mathcal{D}' : A(o) * A(o') \}|,
			% : ->  such that
	%\end{equation}
	\begin{equation}\label{eqn:support}
		sup(GP) = \frac{1}{|\mathcal{D}'|}\cdot |\{(o,o')\in \mathcal{D}' / A(o) * A(o') \}|,
	\end{equation}
	
	where $* \in \{<,>\}$. Given a user-specified threshold $\sigma$, a GP is said to be \textit{frequent} if:
	
	\begin{equation}\label{eqn:min_support}
		sup(GP) \geq \sigma.
	\end{equation}
	
	For example, in Table~\ref{tab:sample_1b}, the gradual item $\{sessions^{\uparrow}\}$ and the GP $\{age^{\uparrow},$ $ sessions^{\uparrow}\}$ are both respected by 3/6 object pairs (i.e. \{(obj1, obj2), (obj1, obj3), (obj2, obj3)\}). Therefore, the frequency support is 0.5 in both scenarios.
	
	We adopt the following notation in the rest of the paper. Let $X = \{gp_1, gp_2,$ $..., gp_n \}$ denote a set of GPs extracted from a data set $\mathcal{D}$. Any GP in $X$ is denoted as \textit{relevant} if and only if $sup(gp_i) \geq \sigma$, where $i \leq n$. The GP in $X$ that has the highest support is denoted as the \textit{best GP}. An \textit{invalid GP} is one that is composed of 2 or more conflicting gradual items (i.e., $\{age^{\uparrow}, marks^{\downarrow}, age^{\downarrow}\}$).
	
	The remainder of the paper is organized as follows: we review related literature in Section~\ref{sec:review}; we define a lean search space for GP candidates built on top of our numeric encoding in Section~\ref{sec:search_space}; we describe a meta-heuristic approach for the problem of mining gradual patterns in Section~\ref{sec:random_algs}; we present our experimental study in Section~\ref{sec:experiments}; finally, we conclude and give future directions concerning the study in Section~\ref{sec:conclusion}.
	
	\section{Review of Literature}
	\label{sec:review}
	Many approaches for efficiently mining GPs have been proposed by numerous researchers. \cite{Berzal2007} describes an Apriori-based algorithm for extracting gradual dependency from data sets using different classes of fuzzy sets. However, according to \cite{Di-Jorio2008}, this approach is limited to extracting GPs of length 3. Instead of using object pairs to verify GPs, \cite{Di-Jorio2008} proposes an ordered data set as an alternative.
	
	In reality, none of the approaches proposed by \cite{Di-Jorio2008} and \cite{Berzal2007} consider the problem of reducing the number of GP candidates that an algorithm has to verify. In view of Equation~\ref{eqn:support} and \ref{eqn:min_support}, it is easy to confirm that the most computationally intensive task in GP mining involves computing frequency support in order to verify/validate GP candidates by comparing their support values against a user-specified threshold. 
	
	In fact, the higher the number of GP candidates to be validated, the more computational resources required by a GP mining algorithm \cite{Owuor2021oct,Clementin2021,Negrevergne2014,Owuor2021jan}. Meta-heuristic based algorithms may be harnessed to accomplish the task of efficiently searching for best GP candidates without having to evaluate all the candidates in the search space. In this study, we investigate LS, RS, GA, PSO algorithms for this task.
		
		% Introduce RS
	Random search (also known as pure random search) algorithms are stochastic algorithms that select and evaluate candidates from a search space independently, and the objective function has no effect on the strategy used to pick the next candidate (i.e. each candidate is selected out of pure randomness) \cite{Solis1981,Zabinsky2003,Zabinsky2011}. RS is a simple and effective approach to frequent pattern mining problems. Therefore, it is surprising how scarcely it has been applied to the problem of searching for frequent pattern candidates.
	
	% Introduce LS
	Local search (also known as neighborhood search) is a meta-heuristic which tries to find optimal solutions by considering the neighbors of current solutions. In a typical combinatorial optimization problem, a search space is defined by a finite set of feasible candidates, each candidate has a cost, and the goal is to find a candidate with the minimum (or maximum) cost \cite{Franti2000,Hoos2004,Ishibuchi1996,Johnson1988}. In the realm of pattern mining, a research study conducted by \cite{Hossain2014} show that LS algorithms are computationally more efficient than systematic search methods in searching for good quality frequent patterns.
		
	% Introduce GA
	Genetic algorithm is an evolutionary algorithm inspired by natural selection systems. The main operators of GA are selection, crossover and mutation: where, selection operation allows GA to sample the \textit{fittest} candidates from the search space (or population), crossover and mutation operations allow strong GA candidates to mate and produce stronger (or better) candidates \cite{Holland1975,Koza1992,Mirjalili2019}. In the realm of frequent pattern mining, research studies conducted by \cite{Kabir2015} and \cite{Saravanan2014} prove that GA-based approaches surpus their classical counterparts in computational performance.
	
	% Introduce PSO
	Particle swarm optimization is a swarm based optimization technique (originally proposed by \cite{Kennedy1995}) that is inspired by the analogy of social interaction and communication (i.e. fish schooling or bird flocking). PSO simulates the movements of swarms in order to iteratively optimize a combinatorial optimization problem \cite{Rajamohana2018}. In the realm of frequent pattern mining, research studies conducted by \cite{Shruti2011} and \cite{Shruti2012} demonstrate how PSO-based approaches improve the performance of the frequent pattern (FP)-growth technique.

	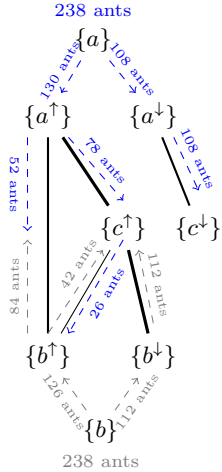
\begin{figure}[h!]
	\centering
	\small
		\begin{tikzpicture}
		
		\begin{scope}[circlestyle/.style={cycle, thin, draw}]%[dotstyle/.style={circle, fill, minimum size=1mm, inner sep=0}]
			\node (nu) at (-0.6,1) {$\{a\}$};
			\node[above, blue] at (-0.6, 1.2) {\scriptsize 238 ants};
			
			\node (n11) at (-1.2,0) {$\{a^{\uparrow}\}$};
			\node (n12) at (0.2,0) {$\{a^{\downarrow}\}$};

			\node (n21) at (-0.2,-1.5) {$\{c^{\uparrow}\}$};
			\node (n22) at (0.8,-1.5) {$\{c^{\downarrow}\}$};
			
			\node (n31) at (-1.2,-3.2) {$\{b^{\uparrow}\}$};
			\node (n32) at (0.2,-3.2) {$\{b^{\downarrow}\}$};
			
			\node (nd) at (-0.5,-4.2) {$\{b\}$};
			\node[above, gray] at (-0.5,-4.8) {\scriptsize 238 ants};
		\end{scope}
	
		\begin{scope}
			\path (nu) edge[draw=blue, dashed, ->] node[blue, sloped, above]{\tiny 130 ants} (n11);
			\path (nu) edge[draw=blue, dashed, ->] node[blue, sloped, above]{\tiny 108 ants} (n12);
			
			\draw[thick] (n11) edge (n31);
			\draw[very thick] (n11) edge (n21);
			\draw[thick] (n12) edge (n22);
			\draw[] (n21) edge (n31);
			\draw[very thick] (n21) edge (n32);
			
			\path (nd) edge[draw=gray, dashed, ->] node[gray, sloped, below]{\tiny 126 ants} (n31);
			\path (nd) edge[draw=gray, dashed, ->] node[gray, sloped, below]{\tiny 112 ants} (n32);
			
			\draw[dashed, blue] ([xshift=-2ex]n11.south) edge[->]  node[sloped, below, blue] {\tiny 52 ants} ([yshift=10ex, xshift=-2ex]n31.north);
			\draw[dashed, blue] ([xshift=2ex]n11.south) edge[->]  node[sloped, above, blue] {\tiny 78 ants} ([yshift=0.5ex]n21.north);
			\draw[dashed, blue] ([xshift=2ex]n12.south) edge[->]  node[sloped, above, blue] {\tiny 108 ants} ([yshift=1.5ex]n22.north);
			
			\draw[dashed, blue] ([yshift=1ex]n21.south) edge[->]  node[sloped, below, blue] {\tiny 26 ants} ([xshift=2ex]n31.north);

			\draw[dashed, gray] ([xshift=-2ex]n31.north) edge[->]  node[sloped, above, gray] {\tiny 84 ants} ([yshift=-10ex, xshift=-2ex]n11.south);
			\draw[dashed, gray] ([yshift=0ex]n31.north) edge[->]  node[sloped, above, near end, gray] {\tiny 42 ants} ([xshift=-2ex]n21.south);
			\draw[dashed, gray] ([yshift=1ex]n32.north) edge[->]  node[sloped, above, gray] {\tiny 112 ants} ([xshift=1.5ex]n21.south);
		\end{scope}		
		\end{tikzpicture}
	\caption{An example of artificial ants for generating GP candidates. Assume that at every iteration there are 2 groups of ants: 238 blue ants arriving at node $\{a\}$ and another 238 gray ants arriving at node $\{b\}$ and, each group of ants try to travel to the opposite node. Recall that every ant in each group is attracted to the path with the highest intensity of pheromone. Initially, the ants select paths haphazardly; but, after a few iterations ants that select shortest paths (illustrated as darker) reach their destinations faster while depositing pheromones. Therefore, ants coming in the opposite direction will find these paths more attractive than the longer paths.}
	\label{fig:artificial_ants}
	\end{figure}
	
	In the realm of GP mining, a research study conducted by \cite{Owuor2021jul} proposes and describes an approach that applies an ant colony optimization (ACO) approach to efficiently search for GP candidates whose frequency support values exceed a user-specified threshold value. According to \cite{Dorigo1996,Dorigo2019,Runkler2005,Stutzle2000}, ACO-based approaches imitate the positive feedback reinforcement behavior of ants as they search for food to solve combinatorial optimization problems.
	
	In \cite{Owuor2021jul}, an ACO variant called \textit{`max-min ant system'} (initially proposed by \cite{Stutzle2000}) is applied to the case GP mining. For instance given attributes $\{a,b,c\}$ of a data set, a search space may be represented such that each point is a path that connects gradual items $\{a^{\uparrow}, a^{\downarrow},b^{\uparrow}, b^{\downarrow},c^{\uparrow}, c^{\downarrow}\}$ as shown in Figure~\ref{fig:artificial_ants}. Each path represents a GP candidate; therefore, the path that is most travelled by ants becomes the best GP candidate.

	Additionally, the study conducted by \cite{Owuor2021jul} implement a GA-based approach and a PSO-based approach that efficiently search for best GP candidates whose frequency support exceed a user-specified threshold value. For case of GA and PSO, \cite{Owuor2021jul} proposes a bitmap encoding to represent each point of a search space as a combination of bits. For instance given attributes $\{a,b,c\}$ of a data set, a search space may be represented such that each point or candidate is encoded as a $(3\times 2)$ bitmap. Figure~\ref{fig:bitmap_ex} illustrates a bitmap encoding of  GP candidate $\{a^{\uparrow},  b^{\downarrow},c^{\uparrow}\}$.
	
	\begin{figure}[h!]
		\centering
		\small
		\begin{tabular}{|c|c c|}
		\hline
		 & \textcolor{magenta}{$\uparrow$} & \textcolor{magenta}{$\downarrow$}\\
		\hline
		\textcolor{magenta}{$\{c\}$} & 1 & 0\\
		\textcolor{magenta}{$\{b\}$} & 0 & 1\\
		\textcolor{magenta}{$\{a\}$} & 1 & 0\\
		\hline
		\end{tabular}
		\caption{An example of a bitmap encoding for a GP candidate $\{a^{\uparrow},  b^{\downarrow},c^{\uparrow}\}$. Where $\uparrow$ denotes \textit{`increasing'} and $\downarrow$ denotes \textit{`decreasing'}.}
		\label{fig:bitmap_ex}
	\end{figure}

	In the final analysis, \cite{Owuor2021jul} presents experimental results which show that ACO-based, GA-based and PSO-based techniques computationally out-perform traditional techniques in mining for GPs. Unlike the traditional techniques, these meta-heuristic based techniques do not need to evaluate all possible GP candidates in the search space.	
	
	\clearpage

	\section{Constructing a Numeric Search Space for GP Candidates}
	\label{sec:search_space}
	In this section, we propose and describe a numeric encoding for GP candidates in Section~\ref{sec:encoding}; we propose and describe how these encodings may be used to define a search space with lower and upper bounds in Section~\ref{sec:decimal_space}; and, we describe an objective function that computes the fitness (or cost) of any selected GP candidate in Section~\ref{sec:cost_function}.

	\subsection{A Numeric Encoding for GP Candidates}
	\label{sec:encoding}
	Generally, the task of extracting GPs begins with generating GP candidates. This step is followed by that of computing the support of the candidates (using the formulas described in Section~\ref{sec:background}) in order to test if they are frequent or not \cite{Owuor2021aug}. Depending on the number of attributes in a data set, the possible number of GP candidates may grow exponentially. This growth may easily computationally overwhelm those algorithms that test every possible GP candidate \cite{Owuor2021aug,Owuor2021jul,Clementin2021}.
		
	\begin{figure*}[h!]
		\centering
		\tiny
		\begin{tikzpicture}%[every node/.style={rectangle, thin, draw}]%
		\begin{scope}
			\node (n0) at (0,0) {$\{\}$};
			
			\node (n11) at (-4.5,1) {$\{A^{\uparrow}\}$};
			\node (n12) at (-2.5,1) {$\{A^{\downarrow}\}$};
			\node (n13) at (-0.5,1) {$\{S^{\uparrow}\}$};
			\node (n14) at (1.5,1) {$\{S^{\downarrow}\}$};
			\node (n15) at (3.5,1) {$\{M^{\uparrow}\}$};
			%\node (n16) at (5.5,1) {$\{M^{\downarrow}\}$};
			\node (n16) at (5,1) {$\{M^{\downarrow}\}$};
			
			\node (n21) at (-6,2.5) {$\{A^{\uparrow},S^{\uparrow}\}$};
			\node (n22) at (-4.9,2.5) {$\{A^{\uparrow},S^{\downarrow}\}$};
			\node (n23) at (-3.8,2.5) {$\{A^{\uparrow},M^{\uparrow}\}$};
			\node (n24) at (-2.7,2.5) {$\{A^{\uparrow},M^{\downarrow}\}$};
			\node (n25) at (-1.6,2.5) {$\{A^{\downarrow},S^{\uparrow}\}$};
			\node (n26) at (-0.5,2.5) {$\{A^{\downarrow},S^{\downarrow}\}$};
			\node (n27) at (0.6,2.5) {$\{A^{\downarrow},M^{\uparrow}\}$};
			\node (n28) at (1.7,2.5) {$\{A^{\downarrow},M^{\downarrow}\}$};
			\node (n29) at (2.8,2.5) {$\{S^{\uparrow},M^{\uparrow}\}$};
			\node (n210) at (3.9,2.5) {$\{S^{\uparrow},M^{\downarrow}\}$};
			\node (n211) at (5,2.5) {$\{S^{\downarrow},M^{\uparrow}\}$};
			%\node (n212) at (6.1,2.5) {$\{S^{\downarrow},M^{\downarrow}\}$};
			
			\node (n31) at (-5.5,4) {$\{A^{\uparrow},S^{\uparrow},M^{\uparrow}\}$};
			\node (n32) at (-3.9,4) {$\{A^{\uparrow},S^{\uparrow},M^{\downarrow}\}$};
			\node (n33) at (-2.3,4) {$\{A^{\uparrow},S^{\downarrow},M^{\uparrow}\}$};
			\node (n34) at (-0.7,4) {$\{A^{\uparrow},S^{\downarrow},M^{\downarrow}\}$};
			\node (n35) at (0.9,4) {$\{A^{\downarrow},S^{\uparrow},M^{\uparrow}\}$};
			\node (n36) at (2.6,4) {$\{A^{\downarrow},S^{\uparrow},M^{\downarrow}\}$};
			\node (n37) at (4.4,4) {$\{A^{\downarrow},S^{\downarrow},M^{\uparrow}\}$};
			%\node (n38) at (5.7,4) {$\{A^{\downarrow},S^{\downarrow},M^{\downarrow}\}$};

		\end{scope}

		\begin{scope}
			\draw[-] (n0) edge[draw=black, thin] (n11);
			\draw[-] (n0) edge[draw=black, thin] (n12);
			\draw[-] (n0) edge[draw=black, thin] (n13);
			\draw[-] (n0) edge[draw=black, thin] (n14);
			\draw[-] (n0) edge[draw=black, thin] (n15);
			\draw[-] (n0) edge[draw=black, thin] (n16);

			\draw[-] (n11.north) edge (n21.south);
			\draw[-] (n13.north) edge (n21.south);
			
			\draw[-] (n11.north) edge (n22.south);
			\draw[-] (n14.north) edge (n22.south);
			
			\draw[-] (n11.north) edge (n23.south);
			\draw[-] (n15.north) edge (n23.south);
			
			\draw[-] (n11.north) edge (n24.south);
			\draw[-] (n16.north) edge (n24.south);
			
			\draw[-] (n12.north) edge (n25.south);
			\draw[-] (n13.north) edge (n25.south);
			
			\draw[-] (n12.north) edge (n26.south);
			\draw[-] (n14.north) edge (n26.south);
			
			\draw[-] (n12.north) edge (n27.south);
			\draw[-] (n15.north) edge (n27.south);
			
			\draw[-] (n12.north) edge (n28.south);
			\draw[-] (n16.north) edge (n28.south);
			
			\draw[-] (n13.north) edge (n29.south);
			\draw[-] (n15.north) edge (n29.south);
			
			\draw[-] (n13.north) edge (n210.south);
			\draw[-] (n16.north) edge (n210.south);
			
			\draw[-] (n14.north) edge (n211.south);
			\draw[-] (n15.north) edge (n211.south);
			
			%\draw[-] (n14.north) edge (n212.south);
			%\draw[-] (n16.north) edge (n212.south);

			\draw[-] (n21.north) edge (n31.south);
			\draw[-] (n23.north) edge (n31.south);
			
			\draw[-] (n21.north) edge (n32.south);
			\draw[-] (n24.north) edge (n32.south);
			
			\draw[-] (n22.north) edge (n33.south);
			\draw[-] (n23.north) edge (n33.south);
			
			\draw[-] (n22.north) edge (n34.south);
			\draw[-] (n23.north) edge (n34.south);
			
			\draw[-] (n25.north) edge (n35.south);
			\draw[-] (n27.north) edge (n35.south);
			
			\draw[-] (n25.north) edge (n36.south);
			\draw[-] (n28.north) edge (n36.south);
			
			\draw[-] (n26.north) edge (n37.south);
			\draw[-] (n27.north) edge (n37.south);
			
			%\draw[-] (n26.north) edge (n38.south);
			%\draw[-] (n28.north) edge (n38.south);

		\end{scope}	
		\end{tikzpicture}
		\caption{A lattice diagram showing the search space universe for GPs based on the attributes of Table~\ref{tab:sample_1b}. Where A denotes age, S denotes sessions and M denotes marks attributes of the data set in Table~\ref{tab:sample_1b}. Due to space constraints, omit GPs: $\{S^{\downarrow},M^{\downarrow}\}$ and $A^{\downarrow},S^{\downarrow},M^{\downarrow}\}$.}
		\label{fig:lattice}
	\end{figure*}
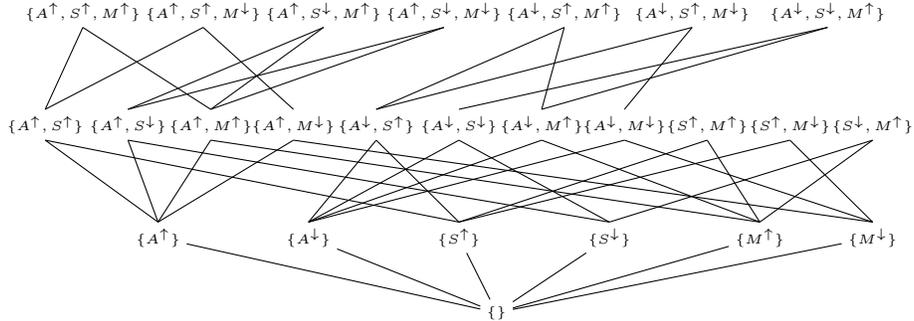
	
	In this paper, we propose and use meta-heuristic approaches (described in Section~\ref{sec:random_algs}) to efficiently learn frequent GP candidates without testing all individuals in the search universe. In this section, we propose and demonstrate a numeric model that effectively encodes all possible GP candidates (of any given data set) in a search universe within an upper and a lower bound.

	To begin with, we mention that a search space for GP candidates is made up of individuals that are formed by arranging a specific set of gradual items in different combinations; and, this set of gradual items is derived from the data set's attribute set. It is important to highlight that the \textit{complementary notion} of GPs permits the existence of 2 gradual items for every attribute \cite{Clementin2021}. 
	
	For instance, the data set in Table~\ref{tab:sample_1b} with the attribute set \{age, sessions, marks\}, allows for 6 gradual items to exist: $\{age^{\uparrow}, age^{\downarrow}, sessions^{\uparrow}, sessions^{\downarrow},$ $marks^{\uparrow}, marks^{\downarrow}\}$. And, Figure~\ref{fig:lattice} is a lattice diagram (or search space) of all possible candidates that may be formed from these 6 gradual items.
	
	Different from the lattice diagram representation, we propose a numeric model that may be used to encode each GP candidate in the search universe as a \textit{bit vector (or bit array/string or bit set)} as illustrated in Table~\ref{tab:bit_encodings}.
			
	\begin{table}[h!]
		\caption{Gradual bit vectors of GP candidates in a search universe. Where A denotes age, S denotes sessions and M denotes marks attributes of the data set in Table~\ref{tab:sample_1b}.}
		\label{tab:bit_encodings}
		\small
		\centering
		\subfloat[]{%
			\begin{tabular}{|c l|}
				\hline
				\textbf{Bit vector} & \textbf{GP}\\
				$(b_{gp})$ & \textbf{Candidate}\\
				\hline
				101000 & $\{A^{\uparrow}, S^{\uparrow}\}$\\
				100100 & $\{A^{\uparrow}, S^{\downarrow}\}$\\
				100010 & $\{A^{\uparrow}, M^{\uparrow}\}$\\
				100001 & $\{A^{\uparrow}, M^{\downarrow}\}$\\
				011000 & $\{A^{\downarrow}, S^{\uparrow}\}$\\
				010100 & $\{A^{\downarrow}, S^{\downarrow}\}$\\
				010010 & $\{A^{\downarrow}, M^{\uparrow}\}$\\
				010001 & $\{A^{\downarrow}, M^{\downarrow}\}$\\
				001010 & $\{S^{\uparrow}, M^{\uparrow}\}$\\
				001001 & $\{S^{\uparrow}, M^{\downarrow}\}$\\
				\hline
			\end{tabular}
		}
		\subfloat[]{%
			\begin{tabular}{|c l|}
				\hline
				\textbf{Bit vector} & \textbf{GP}\\
				$(b_{gp})$ & \textbf{Candidate}\\
				\hline
				000110 & $\{S^{\downarrow}, M^{\uparrow}\}$\\
				000101 & $\{S^{\downarrow}, M^{\downarrow}\}$\\
				101010 & $\{A^{\uparrow}, S^{\uparrow}, M^{\uparrow}\}$\\
				101001 & $\{A^{\uparrow}, S^{\uparrow}, M^{\downarrow}\}$\\
				100110 & $\{A^{\uparrow}, S^{\downarrow}, M^{\uparrow}\}$\\
				100101 & $\{A^{\uparrow}, S^{\downarrow}, M^{\downarrow}\}$\\
				011010 & $\{A^{\downarrow}, S^{\uparrow}, M^{\uparrow}\}$\\
				011001 & $\{A^{\downarrow}, S^{\uparrow}, M^{\downarrow}\}$\\
				010110 & $\{A^{\downarrow}, S^{\downarrow}, M^{\uparrow}\}$\\
				010101 & $\{A^{\downarrow}, S^{\downarrow}, M^{\downarrow}\}$\\
				\hline
			\end{tabular}
		}
	\end{table}
	
	To put it another way, it is possible to compute the length of a bit vector whose individual bits may be toggled in order to provide different encodings for different GP candidates in the search universe, as illustrated in Definition 3.1.
		
	\textbf{Definition 3.1.} \texttt{(Gradual Bit Vector).} \textit{A bit array $b_{gp}$ of size $k$ that encodes a GP candidate; such that $k = (2\cdot |AT|)$, where $AT$ is the attribute set of a data set.}

	For example, we derive a set of 6 gradual items from the attribute set $\{age, sessions, marks\}$ of the data set in Table~\ref{tab:sample_1b} (i.e. $\{age^{\uparrow}, age^{\downarrow}, sessions^{\uparrow},$ $ sessions^{\downarrow},$ $marks^{\uparrow}, marks^{\downarrow}\}$). From this, we construct a gradual bit vector of size 6 (i.e 111111) where each binary digit represents the \textit{position} of each gradual item and if the gradual item is \textit{present} or \textit{absent}. 

	For instance, if items $\{age^{\uparrow}\}$ and $\{marks^{\uparrow}\}$ are the only gradual items present in a GP candidate, then the gradual bit vector may be updated as 100010. Using this numeric model, it is possible to represent all GP candidates in the search universe as gradual bit vectors (as shown in Table~\ref{tab:bit_encodings}).

	\subsection{Defining a Search Space Based on Numeric Encoding}
	\label{sec:decimal_space}
	With attention to the numeric encoding (described in Definition 3.1), it is easy to observe that gradual bit vectors may also represent binary numbers (in the base-2 numeral/positional system) within a specific range \cite{Harris2010,Regan2018}. A lower bound and an upper bound for this range may easily be defined by the minimum and the maximum bit vectors respectively. 
	
	For example, the lower and upper bounds of gradual bit vectors of size 6 may be defined by the values 000000 and 111111 respectively. Notably, (in Table~\ref{tab:bit_encodings}) all gradual bit vectors are located within the bounds of these two values.
	
	A part from the binary numeral system, there exist numerous positional numeral systems that may be used in computational arithmetic. The conventional decimal numeral system (which uses 10 digits 0, 1, .., 9) is one of most popular number systems in use among various research domains \cite{Regan2018}. We can convert each gradual bit vector into its decimal value and hexadecimal value equivalent. 
		
	\begin{table}[h!]
		\caption{Decimal values of bit vector encodings obtained from Table~\ref{tab:bit_encodings}.}
		\label{tab:dec_encodings}
		\small
		\centering
		\subfloat[]{%
			\begin{tabular}{|l c c|}
				\hline
				\textbf{Bit} & \textbf{Decimal}& \textbf{Hexa}\\
				\textbf{Vector} & \textbf{Value} & \textbf{Value}\\
				\hline
				101000 & 40 & 28\\
				100100 & 36 & 24\\
				100010 & 34 & 22\\
				100001 & 33 & 21\\
				011000 & 24 & 18\\
				010100 & 20 & 14\\
				010010 & 18 & 12\\
				010001 & 17 & 11\\
				001010 & 10 & A\\
				001001 & 09 & 09\\
				\hline
			\end{tabular}
		}
		\subfloat[]{%
			\begin{tabular}{|l c c|}
				\hline
				\textbf{Bit} & \textbf{Decimal} & \textbf{Hexa}\\
				\textbf{Vector} & \textbf{Value} & \textbf{Value}\\
				\hline
				000110 & 06 & 06\\
				000101 & 05 & 05\\
				101010 & 42 & 2A\\
				101001 & 41 & 29\\
				100110 & 38 & 26\\
				100101 & 37 & 25\\
				011010 & 26 & 1A\\
				011001 & 25 & 19\\
				010110 & 22 & 16\\
				010101 & 21 & 15\\
				\hline
			\end{tabular}
		}
	\end{table}
	
	%\newpage
	For example, the bit vectors in  Table~\ref{tab:bit_encodings} may be converted into their decimal value equivalents as shown in Table~\ref{tab:dec_encodings}. We observe that all positions of the decimal values (that represent the gradual bit vector) lie within the bounds of 5 and 42 (which are the lower and the upper bound values respectively). Therefore, we may define a GP search space based on numeric encoding as:
	
	\noindent \textit{``an interval $\mathcal{S}_{gp}$ of integer numbers within the bounds of $a$ and $b$ such that, for any $x \in \mathcal{S}_{gp}$, $a \leq x \leq b$''.}
	
	Our numeric encoding simplifies the computation of determining the values of bounds $a$ and $b$ respectively. In the case of lower bound $a$: it is easy to verify that for any data set with at least 2 attributes, the smallest decimal value that corresponds to valid bit vector is 5. For this reason, it is true that $a = 5$ always.
	
	In the case of upper bound $b$, we propose a different approach for computing its value based on the size of the attribute set of a data set. Using Table~\ref{tab:dec_encodings}, we observe that the largest value of a bit vector representing a 2-attribute data set is 1010, and that of a 3-attribute data set is 101010, and that of a 4-attribute data set is 10101010, and so forth. Therefore, $b$ can be computed from the sum of odd binary digits times their power of 2 $(2^{n})$ as shown in Equation~\eqref{eqn:max_bound}.
	
	\begin{equation}\label{eqn:max_bound}
		b = \sum \limits _{i=1} ^{|AT|} 2^{2i - 1},
	\end{equation}
	
	where $|AT|$ is the number of attributes. Therefore, we may define $\mathcal{S}_{gp}$ as:
	%Therefore, we may define our search space using the following formula:
	
	\begin{equation}
		\mathcal{S}_{gp} = \Big\{x \in \mathbb{Z} ~\Big|~ 5 \leq x \leq \textstyle\sum _{i=1}^{|AT|} 2^{2i - 1} \Big\}.
		%\Big\{x ~\Big|~ x\in [5, \textstyle\sum _{i=1}^{|AT|} 2^{2i - 1}] \Big\}.
	\end{equation}
	
	It is interesting to note that the search space proposed in \cite{Owuor2021jul} is based on $[|AT| \times 2]$ binary encoding. Therefore, its dimensions are huge compared to our proposed search space. For example, given 3 attributes, the search space based on binary encoding allows algorithms to traverse almost $(2^6 = 64)$ possible candidate combinations, while our proposed search space allows algorithms to traverse only $(42-5=37)$ candidates. The numeric encoding reduces the search space by almost half of that which was proposed in \cite{Owuor2021jul}.
	
	However, not all gradual bit vectors correspond to valid GP candidates. For example, using the data set in Table~\ref{tab:sample_1b}, 001111 is an invalid GP candidate since it represents conflicting gradual items (i.e., $\{sessions^{\uparrow}, sessions^{\uparrow},$ $marks^{\uparrow}, marks^{\uparrow}\}$). In such scenarios, GP algorithms ignore such candidates.
	
	Our proposed numeric search space (although to a lesser amount than the binary search space) still includes some invalid GPs. Using the lattice diagram in Figure~\ref{fig:lattice}, we learn that a data set with 3 attributes yields only 20 valid GP candidates. Therefore, our search space contains 17 invalid GP candidates that should be ignored by mining algorithms.

	\subsection{Computing Candidates' Fitness}
	\label{sec:cost_function}
	Each gradual bit vector should be associated with a fitness/cost value. The fitness value should allow these algorithms to learn (using a probabilistic search) which gradual bit vector encodes the GP candidate with the highest frequency support (described in Section~\ref{sec:background}).
	
	In this paper, we state that the gradual bit vector with the smallest fitness (or cost) value is the best candidate. Given this point, we observe that there exists a relationship between frequency support of a candidate and fitness. Therefore, a slight modification of Equation~\ref{eqn:support} may yield an objective function for determining the fitness of GP candidates. Therefore, the objective function may be defined by the inverse of the concordant object pairs count as shown in Equation~\ref{eqn:fitness_func}.
	
	\begin{equation}\label{eqn:fitness_func}
		fitness(b_{gp}) = \frac{1}{|\{\forall(o,o')\in \mathcal{D}' : A(o) * A(o') \}|}.
	\end{equation}
		
	For example, given the data set in Table~\ref{tab:sample_1b}, we construct some gradual bit vectors and compute their decimal value equivalents and their fitness values as shown in Table~\ref{tab:cost_values}.
	
	\begin{table}[h!]
		\caption{A table illustrating the association between bit vectors, GP candidates, decimal values and fitness values respectively. Where A denotes age, S denotes sessions and M denotes marks attributes of the data set in Table~\ref{tab:sample_1b}.}
		\label{tab:cost_values}
		\small
		\centering
			\begin{tabular}{|l l c c|}
				\hline
				\textbf{Bit} & \textbf{GP} & \textbf{Decimal} & \textbf{Fitness}\\
				\textbf{vector} & \textbf{Candidate} & \textbf{value} & \textbf{value}\\
				\hline
				101000 & $\{A^{\uparrow}, S^{\uparrow}\}$ & 40 & 0.25\\
				100001 & $\{A^{\uparrow}, M^{\downarrow}\}$ & 33 & 0.5\\
				011010 & $\{A^{\downarrow}, S^{\uparrow}, M^{\uparrow}\}$ & 26 & 1.0\\
				\hline
			\end{tabular}
	\end{table}
	
	Using Table~\ref{tab:sample_1b}, the pattern $\{A^{\uparrow}, S^{\uparrow}\}$ is respected by 4 out of 6 object pairs: \{(obj1, obj2), (obj1, obj3), (obj1, obj4), (obj2, obj3)\}. Therefore, the support is 0.67 (see Equation~\ref{eqn:support}) and the fitness value is 0.25 (see Equation~\ref{eqn:fitness_func}). As shown in Table~\ref{tab:cost_values}, the GP $\{A^{\uparrow}, S^{\uparrow}\}$ has the lowest fitness value and highest support in comparison to GPs $\{A^{\uparrow}, M^{\downarrow}\}$ and $\{A^{\downarrow}, S^{\uparrow}, M^{\uparrow}\}$.

	\section{Metaheuristic Algorithms for GP Extraction}
	\label{sec:random_algs}
	In this section, we propose 4 meta-heuristic algorithms (i.e., LS, RS, GA, PSO) to the problem of finding best GP from the search space proposed in Section~\ref{sec:cost_function}.

	\subsection{LS-GRAD and RS-GRAD Algorithms}
	\label{sec:ls_rs}
	We propose RS-GRAD algorithm for GP extraction and it is based on a \textit{random search algorithm} which is a technique whose optimization strategy is based on a purely stochastic process. In every iteration, random search algorithms are designed to modify the current solution by a random factor \cite{Zabinsky2003}.
	
	Similarly, we propose LS-GRAD algorithm for GP extraction and it is based on a \textit{stochastic hill climbing algorithm}. Hill climbing technique is an iterative technique that starts from an arbitrary solution and attempts to find a better neighboring solution. A hyper-parameter known as \textit{`step-size'} controls how far the search for a neighboring solution is allowed to go \cite{Nolle2006}.
	
	The main steps of LS-GRAD algorithm and RS-GRAD algorithm are illustrated in  Algs. 1(a) and 1(b) respectively.
	
	\begin{figure}[h!]
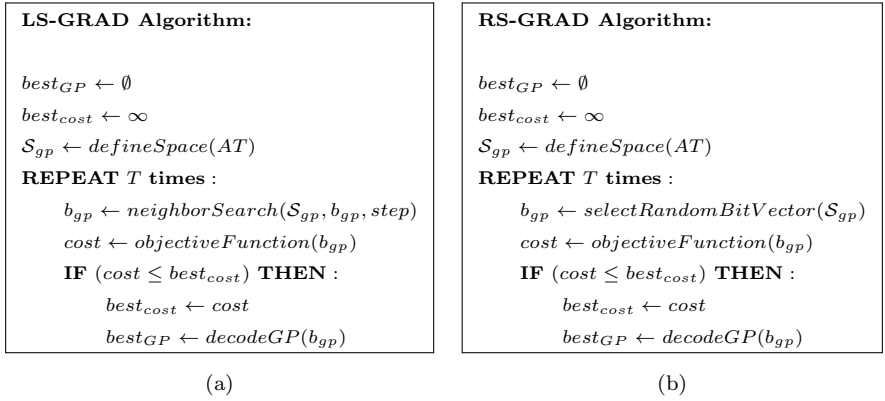

		\centering
		\scriptsize
		% First 
		\subfloat[]{%
			\begin{tabular}{|l|}
				\hline
				\textbf{LS-GRAD Algorithm:} \\
				\\
				$best_{GP} \leftarrow \emptyset$\\
				$best_{cost} \leftarrow \infty$\\
				$\mathcal{S}_{gp} \leftarrow defineSpace(AT)$\\
				$\mathbf{REPEAT}~ T ~\mathbf{times:}$\\
				$\qquad b_{gp} \leftarrow neighborSearch(\mathcal{S}_{gp},b_{gp},step)$\\
				$\qquad cost \leftarrow objectiveFunction(b_{gp})$\\
				$\qquad \mathbf{IF}~ (cost \leq best_{cost}) ~\mathbf{THEN:}$\\
				$\qquad \qquad best_{cost} \leftarrow cost$\\
				$\qquad \qquad best_{GP} \leftarrow decodeGP(b_{gp})$\\
				\hline
			\end{tabular}}
			$~~~$
		% Second
		\subfloat[]{%
			\begin{tabular}{|l|}
				\hline
				\textbf{RS-GRAD Algorithm:} \\
				\\
				$best_{GP} \leftarrow \emptyset$\\
				$best_{cost} \leftarrow \infty$\\
				$\mathcal{S}_{gp} \leftarrow defineSpace(AT)$\\
				$\mathbf{REPEAT}~ T ~\mathbf{times:}$\\
				$\qquad b_{gp} \leftarrow selectRandomBitVector(\mathcal{S}_{gp})$\\
				$\qquad cost \leftarrow objectiveFunction(b_{gp})$\\
				$\qquad \mathbf{IF}~ (cost \leq best_{cost}) ~\mathbf{THEN:}$\\
				$\qquad \qquad best_{cost} \leftarrow cost$\\
				$\qquad \qquad best_{GP} \leftarrow decodeGP(b_{gp})$\\
				\hline
			\end{tabular}}
		\caption*{Alg. 1: (a) LS-GRAD algorithm, (b) RS-GRAD algorithm.}
		\label{fig:sto_algs}
	\end{figure}
	
	As shown in Alg. 1(a), LS-GRAD algorithm defines a search space $S_{gp}$ such that each point is encoded as numeric digit that represents a GP candidate and it iterates $T$ times. In each iteration, the algorithm:
	\begin{itemize}
		\item finds a neighbor of the current selected bit vector candidate $b_{gp}$ in search space $S_{gp}$ and this neighbor becomes the selected candidate (using $neighbor$ $Search()$ function);
		\item evaluates fitness of the selected candidate (using $objectiveFunction()$).
	\end{itemize}
	
	As shown in Alg. 1(b), RS-GRAD algorithm defines a search space $S_{gp}$ such that each point is encoded as numeric digit that represents a GP candidate and it iterates $T$ times. In each iteration, the algorithm:
	\begin{itemize}
		\item randomly selects a bit vector candidate from search space $S_{gp}$ (using $selectRandomBitVector()$ function);
		\item evaluates fitness of the selected candidate (using $objectiveFunction()$).
	\end{itemize}

	\subsection*{Computational Complexity of LS-GRAD and RS-GRAD}
	\label{sec:ls_rs_complexity}
	We use big-O notation \cite{Cormen2009,Vaz2017} to analyze the computational complexity of LS-GRAD and RS-GRAD algorithms. For every candidate that is selected, both LS-GRAD and RS-GRAD algorithms compute fitness using the $objective$ $Function()$ (or fitness function described in Section~\ref{sec:cost_function}). We recall that the formula proposed for computing the fitness function (see Equation~\ref{eqn:fitness_func}) is a variant of the formula used for computing frequency support (see Equation~\ref{eqn:support}). Therefore, the computational complexities of implementing these two functions are fairly similar. According to \cite{Owuor2021jul}, given a data set with $a$ attributes and $n$ objects, the task of computing the fitness of a GP candidate composed of $k$ gradual items (where $k\leq a$) has a complexity of $O(k\cdot n^{2}$).
	
	Another key point is that in both LS-GRAD and RS-GRAD algorithms, the tasks of implementing functions like $defineSpace()$, $decodeGP()$, $neighbor$ $Search()$ and $selectRandomBitVector()$ have small and almost constant computational complexities relative to the task of the $objectiveFunction()$. Therefore, it is acceptable to simplify the big-O analysis to focusing on computationally intensive tasks (i.e. $objectiveFunction()$). Given these points, if both LS-GRAD and RS-GRAD algorithms iterate $T$ times, their computational complexities are relatively of the same equivalence - that is $O(T\cdot k\cdot n^{2})$.

	\subsection{GA-GRAD and PSO-GRAD Algorithms}
	\label{sec:ga_pso}
	We propose GA-GRAD algorithm for GP extraction and it is built on top of GA which is an optimization technique that is inspired by Darwin's theory about evolution \cite{Obitko1999}. GA techniques are initiated with a population of solutions (also called \textit{chromosomes}). Existing solutions are selected according to their fitness and they mate to reproduce offsprings whose fitnesses are better.  
	
	Through this reproduction operation GA techniques efficiently find best solutions to combinatorial optimization problems. Specifically, GA uses the following operators in order to improve the fitness of the offsprings:
	\begin{itemize}
		\item Crossover: passing of good genes to next generation. $\gamma$ denotes the parameter that controls crossover rate.
		\item Mutation: random changes in genes during crossover. $\mu$ denotes the parameter that controls mutation rate.
	\end{itemize}
		
	Similarly we propose PSO-GRAD algorithm for GP extraction and it is built on top of PSO which is an optimization technique that is inspired by collective behavior and movements of swarms (especially as they search for food). PSO techniques are initialized with a population of particles and in every iteration each particle moves to a new position whose cost/fitness is better than its current position \cite{Shruti2011}. 
	
	\textcolor{brown}{A PSO algorithm performs a constant search for best position in a search space. Particles are generated and they are randomly assigned positions from a uniform distribution with the search space; the algorithm moves particles with a certain calculated velocity in every iteration. Personal best position ($p_{i}$ and global best position ($gbest$) are the best position visited so-far by the $i$-th particle and the best position visited so far by any particle in the swarm \cite{Piotrowski2020}. PSO uses the following operators:}
	
	\begin{itemize}
		\item Current direction/velocity: determines the movement of each particle.
		\item Personal best position ($p_{i}$) and cognitive component ($coef_{p}$): $coef_{p}$ co-efficient gives the importance of personal best value.
		\item Team/global best position ($gbest$) and social component ($coef_{g}$): $coef_{g}$ co-efficient gives the importance of global best value.
	\end{itemize}

	The main steps of GA-GRAD algorithm and PSO-GRAD algorithm are illustrated in Algs. 2(a) and 2(b) respectively.
	
	\clearpage
	\begin{figure}[h!]
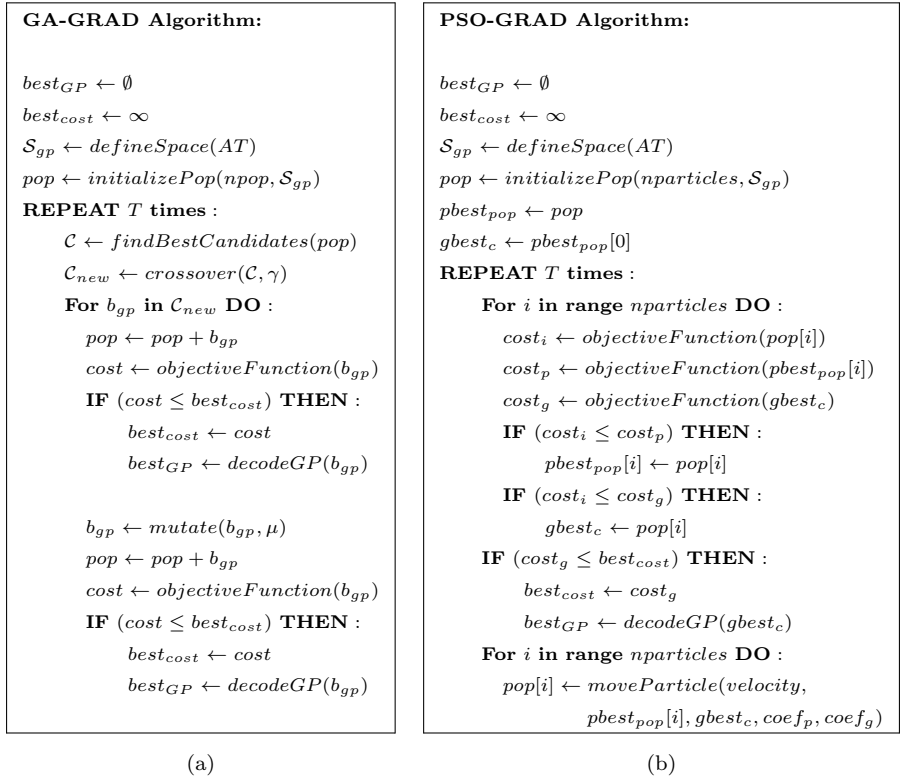

		\centering
		\scriptsize
		% First 
		\subfloat[]{%
			\begin{tabular}{|l|}
				\hline
				\textbf{GA-GRAD Algorithm:} \\
				\\
				$best_{GP} \leftarrow \emptyset$\\
				$best_{cost} \leftarrow \infty$\\
				$\mathcal{S}_{gp} \leftarrow defineSpace(AT)$\\
				$pop \leftarrow initializePop(npop,\mathcal{S}_{gp})$\\
				
				$\mathbf{REPEAT}~ T ~\mathbf{times:}$\\
				$\qquad \mathcal{C} \leftarrow findBestCandidates(pop)$\\
				$\qquad \mathcal{C}_{new} \leftarrow crossover(\mathcal{C},\gamma)$\\	
								
				$\qquad \mathbf{For}~ b_{gp}~ \mathbf{in}~ \mathcal{C}_{new}~ \mathbf{DO:}$\\
				$\qquad \quad pop \leftarrow pop + b_{gp}$\\
				$\qquad \quad cost \leftarrow objectiveFunction(b_{gp})$\\
				$\qquad \quad \mathbf{IF}~ (cost \leq best_{cost}) ~\mathbf{THEN:}$\\
				$\qquad \quad \qquad best_{cost} \leftarrow cost$\\
				$\qquad \quad \qquad best_{GP} \leftarrow decodeGP(b_{gp})$\\
				
				$~$\\
				$\qquad \quad b_{gp} \leftarrow mutate(b_{gp},\mu)$\\
				$\qquad \quad pop \leftarrow pop + b_{gp}$\\

				$\qquad \quad cost \leftarrow objectiveFunction(b_{gp})$\\
				$\qquad \quad \mathbf{IF}~ (cost \leq best_{cost}) ~\mathbf{THEN:}$\\
				$\qquad \quad \qquad best_{cost} \leftarrow cost$\\
				$\qquad \quad \qquad best_{GP} \leftarrow decodeGP(b_{gp})$\\
				$~$\\
				
				\hline
			\end{tabular}}
			$~~~$
		% Second
		\subfloat[]{%
			\begin{tabular}{|l|}
				\hline
				\textbf{PSO-GRAD Algorithm:} \\
				\\
				$best_{GP} \leftarrow \emptyset$\\
				$best_{cost} \leftarrow \infty$\\
				$\mathcal{S}_{gp} \leftarrow defineSpace(AT)$\\
				$pop \leftarrow initializePop(nparticles,\mathcal{S}_{gp})$\\
				$pbest_{pop} \leftarrow pop$\\
				$gbest_{c} \leftarrow pbest_{pop}[0]$\\
				
				$\mathbf{REPEAT}~ T ~\mathbf{times:}$\\
				$\qquad \mathbf{For}~ i~ \mathbf{in~range}~ nparticles~ \mathbf{DO:}$\\
				$\qquad \quad cost_{i} \leftarrow objectiveFunction(pop[i])$\\
				$\qquad \quad cost_{p} \leftarrow objectiveFunction(pbest_{pop}[i])$\\
				$\qquad \quad cost_{g} \leftarrow objectiveFunction(gbest_{c})$\\
				$\qquad \quad \mathbf{IF}~ (cost_{i} \leq cost_{p}) ~\mathbf{THEN:}$\\
				$\qquad \quad \qquad pbest_{pop}[i] \leftarrow pop[i]$\\
				
				$\qquad \quad \mathbf{IF}~ (cost_{i} \leq cost_{g}) ~\mathbf{THEN:}$\\
				$\qquad \quad \qquad gbest_{c} \leftarrow pop[i]$\\
				
				$\qquad \mathbf{IF}~ (cost_{g} \leq best_{cost}) ~\mathbf{THEN:}$\\
				$\qquad \qquad best_{cost} \leftarrow cost_{g}$\\
				$\qquad \qquad best_{GP} \leftarrow decodeGP(gbest_{c})$\\

				$\qquad \mathbf{For}~ i~ \mathbf{in~range}~ nparticles~ \mathbf{DO:}$\\
				$\qquad \quad pop[i] \leftarrow moveParticle(velocity,$\\
				$\qquad \qquad \qquad \quad pbest_{pop}[i], gbest_{c}, coef_{p}, coef_{g}) $\\

				\hline
			\end{tabular}}
		\caption*{Alg. 2: (a) GA-GRAD algorithm, (b) PSO-GRAD algorithm.}
		\label{fig:sbo_algs}
	\end{figure}
	
	As shown in Alg. 2(a), GA-GRAD algorithm is initialized to a population $pop$ having $npop$ candidates and each candidate is a gradual bit vector $b_{gp}$ within search space $\mathcal{S}_{gp}$. The algorithm iterates $T$ times and in each iteration it: 
	\begin{itemize}
		\item selects 2 of the best candidates from population $pop$ (using $findBest$ $Candidates()$ function);
		\item performs a crossover function using these 2 candidates in order to reproduce 2 offsprings (using $crossover()$ function at a rate of $\gamma$);
		\item evaluates the fitness of the offsprings (using $objectiveFunction()$);
		\item mutates each offspring's bit vector (using $mutate()$ function at a rate of $\mu$) and evaluates their fitness;
		\item adds the offsprings to population $pop$.
	\end{itemize}
	
	As shown in Alg. 2(b), PSO-GRAD algorithm is initialized to a population $pop$ having $n$ particles and each particle's position represents a gradual bit vector within search space $\mathcal{S}_{gp}$. In addition, a global best particle $gbest_{c}$ is initialized. The algorithm iterates $T$ times and in each iteration, it:
	\begin{itemize}
		\item evaluates the fitness of all particles' positions in the population and the fitness of the global best particle's position (using $objectiveFunction()$);
		\item for each particle's position if the fitness value is lower than the fitness value of its personal best position in history, set the personal best position to this position;
		\item replace the global best particle position with any particle position whose fitness value is lower than the laters;
		\item moves each particle by a velocity $vel$ towards its personal best at a rate of $coef_{p}$ and towards the global best particle at a rate of $coef_{g}$. 
	\end{itemize}

	\subsection*{Computational Complexity of GA-GRAD and PSO-GRAD}
	\label{sec:ga_pso_complexity}
	We use big-O notation \cite{Cormen2009,Vaz2017} to analyze the computational complexity of GA-GRAD and PSO-GRAD algorithms. Similar to LS-GRAD and RS-GRAD algorithms, we take the computational complexity of a single $objectiveFunction()$ (implemented by GA-GRAD and PSO-GRAD algorithms) as equivalent to $O(k\cdot n^{2})$. Again, comparable to GA-GRAD and PSO-GRAD algorithms, we ignore functions whose computational complexities are small and almost constant relative to the $objectiveFunction()$.
	
	GA-GRAD algorithm is initialized to a population of $npop$ candidates where only 2 candidates are selected in every iteration. Again, the algorithm evaluates the fitness of 2 crossed offsprings and 2 mutated offsprings in every iteration. Therefore, in 1 iteration the computational complexity of GA-GRAD algorithm is equivalent to $O(4\cdot k\cdot n^{2})$ and in $T$ iterations it is equivalent to $O(4\cdot T\cdot k\cdot n^{2})$ 
	
	PSO-GRAD algorithm is initialized to a population of $n$ particles. In each iteration, the algorithm evaluates the fitness of all these particles together with the fitness of their historical personal best and global best particle. Therefore, in 1 iteration the computational complexity of PSO-GRAD algorithm is equivalent to $O\{(2n + 1)\cdot k\cdot n^{2}\}$ and in $T$ iterations it is equivalent to $O\{(2n + 1)\cdot T\cdot k\cdot n^{2}\}$.

	\section{Experiments}
	\label{sec:experiments}
	In this section, we present an experimental study of the computational performance of our algorithms. Specifically, we analyze the performance and behavior of the algorithms when exploring search spaces based on either a bitmap encoding (initially proposed in \cite{Owuor2021jul}) or a numeric encoding (proposed in Section~\ref{sec:search_space}). The aim of this experiment is to establish which search space is simpler for meta-heuristic algorithms. 
	
	In addition to this, we compare the computational performance of
	 the meta-heuristic algorithms in mining GPs. All experiments were conducted on a HPC (High Performance Computing) platform \textbf{Meso@LR}\footnote{\url{https://meso-lr.umontpellier.fr}}. We used one node made up of 14 cores and 128GB of RAM.
	
	\subsection{Source Code}
	\label{sec:code}
	The Python source code of our algorithms is available at the GitHub repository \url{https://github.com/owuordickson/swarm_gp.git}. The Python package for installing these algorithms is available at the PyPi repository through this link: \url{https://pypi.org/project/so4gp/}.
	
	\subsection{Data Set Description}
	\label{sec:dataset}
	Table~\ref{tab:dataset} briefly describes the features of the data sets used in our experiments for evaluating the performance of our algorithms. All these data sets are numeric data set. We performed a data cleaning task (on all the data sets) that involved omitting objects with one or more \textit{missing values} and omitting attributes whose objects are either \textit{non-numeric values} or \textit{time-stamp values}. All these data sets were retrieved from the UCI Machine Learning Repository\footnote{\url{https://archive.ics.uci.edu/ml/index.php}}\cite{Dua2019}. 
	
	\begin{table}[h!]
	\scriptsize
	\centering
	\caption{Experiment data sets.}
	\begin{tabular}{|l|c|c|l|c|}
		\hline
		Data set & $\#$objects used& $\#$attributes & Domain & Source\\
		\hline \hline
		Breast Cancer (BCR)& 116 & 10 & Medical & \cite{Patricio2018}\\
		Hepatitis C (HCV) & 615 & 14 & Medical & \cite{Hoffmann2018}\\
		Chickenpox (CPX) & 521 & 20 & Zoonosis & \cite{Benedek2021}\\
		Cargo 2000 (C2K) & 3942 & 98 & Transport & \cite{Metzger2015}\\
		Air Quality (AQY) & 9358 & 15 & Environment & \cite{DeVito2008}\\
		APS Data (APS)& 2474 & 171 & Vehicle & \cite{Gondek2016}\\
		Omnidir (OMD)& 2000 & 11 & Coastline & \cite{Bouchette2019}\\
		Directio (DIR)& 8075 & 21 & Coastline & \cite{Bouchette2019}\\
		Power Consumption (HPC)& 10001 & 9 & Energy & \cite{Dua2019}\\
		\hline
	\end{tabular}
	\label{tab:dataset}
	\end{table}

	The HCV data set is a data set that contains laboratory values of blood donors of healthy individuals and individuals infected with Hepatitis C or Fibrosis or Cirrhosis together with demographic values (i.e. age, sex) \cite{Hoffmann2018}. The CPX data set is a spatio-temporal data set containing county-level records of weekly chickenpox cases in Hungary between the years of 2005 and 2015 \cite{Benedek2021}. 
	
	The C2K data set is a data set that holds records of Cargo 2000 airfreight tracking and events for a period of 5 months. The C2K data set has 98 attributes and 3942 objects (some with missing values) \cite{Metzger2015}. The AQY data set is a data set that contains records of a gas multi-sensor device (i.e. hourly averages of gas concentrations) deployed on a field in an Italian city \cite{DeVito2008}.
	
	The BCR data set is composed of 10 predictors and a binary variable indicating the presence or absence of breast cancer. The predictors are recordings of anthropometric data gathered from the routine blood analysis of 116 participants \cite{Patricio2018}. The HPC data set describes the electric power consumption in one household (located in Sceaux, France) between 2006 and 2010 \cite{Dua2019}.
	
	The APS data set holds data of the \textit{Air Pressure System} (APS) collected from heavy Scania trucks in their daily usage \cite{Gondek2016}. The DIR and OMD data sets are obtained from OREMES’s data portal\footnote{\url{https://data.oreme.org}} that recorded swell sensor signals of 4 buoys near the coast of the Languedoc-Roussillon region in France between 2012 and 2019 \cite{Bouchette2019}.
	
	\subsection{Parameter Tuning}
	\label{sec:parameter_tuning}
\begin{figure}[h!]
		\centering
		\scriptsize
		\includegraphics[scale=0.36]{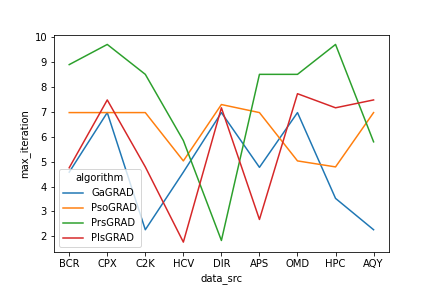}
		\caption{Determining the appropriate maximum value of iterations ($T$). Where GaGRAD denotes GA-GRAD, PsoGRAD denotes PSO-GRAD, PlsGRAD denotes LS-GRAD, PrsGRAD denotes RS-GRAD.}
		\label{fig:exp_tuning1}
	\end{figure}

	In order to obtain the highest efficiency from GA-GRAD, PSO-GRAD and LS-GRAD algorithms, careful tuning of their respective parameters is required. We utilize a Bayesian optimization approach proposed by \cite{Snoek2012} to achieve this task. Firstly, we determine the appropriate maximum value of iterations $T$ (see Section~\ref{sec:random_algs}) for all the algorithms when applied to all the data sets described in Section~\ref{sec:dataset}. As can be seen in Figure~\ref{fig:exp_tuning1}, we observe that none of the algorithms exceeds 10 iterations when applied on any data set. Therefore, we set the maximum value of iterations to 20 in order to how smoothly each algorithm's plot settles in Section~\ref{sec:results}.
	
\begin{figure}[h!]
		\centering
		\scriptsize
		% First 
		\subfloat[]{%
		\includegraphics[scale=0.32]{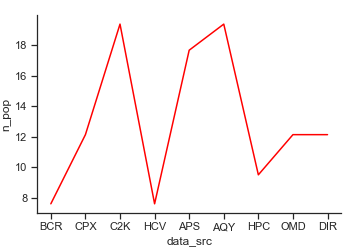}
		}
		% Second 
		\subfloat[]{%
		\includegraphics[scale=0.32]{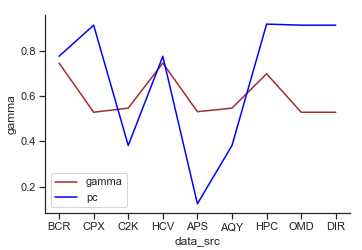}
		}
		% Third
		\subfloat[]{%
		\includegraphics[scale=0.32]{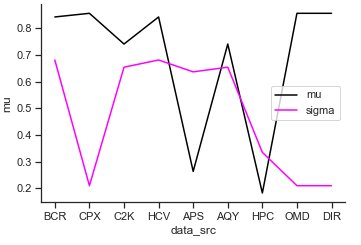}
		}
		\caption{GA-GRAD parameters: (a) initial parent population, (b) offspring proportion (pc) and crossover $\gamma$ parameter, (c) mutation $\mu$ and $\sigma$ parameters.}
		\label{fig:exp_tuning_ga}
	\end{figure}

\begin{figure}[h!]
		\centering
		\scriptsize
		% First 
		\subfloat[]{%
		\includegraphics[scale=0.32]{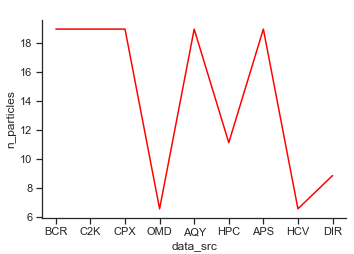}
		}
		% Second 
		\subfloat[]{%
		\includegraphics[scale=0.32]{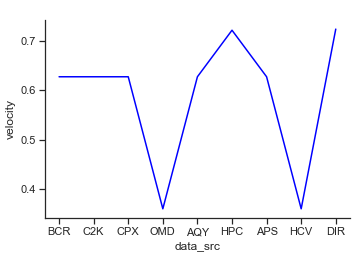}
		}
		% Third 
		\subfloat[]{%
		\includegraphics[scale=0.32]{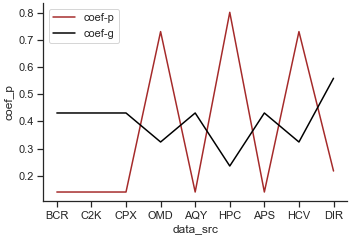}
		}
		\caption{PSO-GRAD parameters: (a) initial particle population, (b) velocity parameter, (c) personal and global coefficients.}
		\label{fig:exp_tuning_pso}
	\end{figure}

\begin{figure}[h!]
		\centering
		\scriptsize
		\includegraphics[scale=0.32]{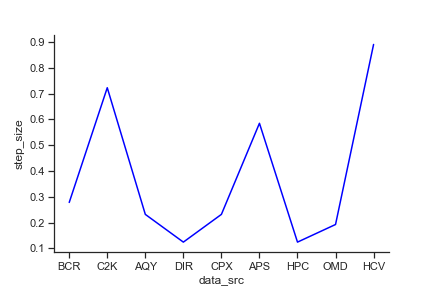}
		\caption{LS-GRAD's step-size parameter.}
		\label{fig:exp_tuning_pls}
	\end{figure}

	Secondly, for each algorithm we determine the optimum values of their respective parameters when applied on different data sets by using the automated Bayesian optimization approach \cite{Snoek2012}. Figures~\ref{fig:exp_tuning_ga}, \ref{fig:exp_tuning_pso}, \ref{fig:exp_tuning_pls} show a plot of specific parameters against different data sets for GA-GRAD, PSO-GRAD and LS-GRAD algorithms respectively. From results displayed in these figures, it is can be deduced that each algorithm needs a different combination of parameter values in order to perform optimally for different data sets.
		
	For this reason, in order to reach maximum the efficiency of each our algorithms in our experimental analysis presented in Section~\ref{sec:results}, we tune their respective parameters to different values according to the data set involved.
		
	\subsection{Experiment Results}
	\label{sec:results}
	In this section, we present our experiment results which reveal the computational behavior of LS-GRAD, RS-GRAD, GA-GRAD, PSO-GRAD algorithms when applied to the data sets described in Section~\ref{sec:dataset}. All experiments involve performing at least 3 repeated test runs on each algorithm and the results are a mean of the repeated test runs of each algorithm on each data set. A single test run allows each algorithm to execute 20 iterations ($T = 20$) in each search space. In every iteration, all the algorithms try to find the best GP candidate among those provided by each search space.
	
	In Section~\ref{sec:exp_wilcoxon}, we perform a \textit{Wilcoxon test} to determine which search space better suites our algorithms. We conduct 2 experiments on the 4 algorithms: one using bitmap encoding to construct search spaces in and, another using numeric encoding to construct search spaces.
	
	In Section~\ref{sec:exp_scatter}, we compare the distribution of \textit{valid} GPs and \textit{invalid} GPs extracted from various data sets by our algorithms using the numeric-based search space against the bitmap-based search space. In Section~\ref{sec:exp_compare}, we conduct an experiment that compares the performance of our algorithms using the numeric-based search space against other existing GP mining algorithms

	\subsubsection{Experiment 1: Wilcoxon Test on Search Spaces}
	\label{sec:exp_wilcoxon}
	The Wilcoxon test is a non-parametric measure that compares the differences between at most 2 paired groups of data \cite{Carrasco2020}. In this study, we evaluate the effect of different search spaces (i.e., numeric-based and bitmap-based) on our algorithms' run-time durations in different data sets. The dependent variable is the run-time duration measured on a continuous scale. The Wilcoxon test uses the following null and alternative hypotheses:
	
	\begin{itemize}
		\item Null hypothesis ($H_{0}$): the mean run-time for each group is equal.
		\item Alternative hypothesis: ($H_{a}$): one group's mean run-time is different from the other.
	\end{itemize}

	\begin{table}[h!]
  	\centering
    \scriptsize
   	\caption{Summary of run-times of algorithms on different data sets. Where Ga denotes GA-GRAD Pso denotes PSO-GRAD, Prs denotes RS-GRAD, Pls denotes LS-GRAD algorithms respectively and NU denotes the numeric-based search space, BM denotes the bitmap-based search space. For instance Ga-NU implies GA-GRAD using the numeric-based search space.}
    	\begin{tabular}{l | c c c c c c c c}
    	\multirow{2}{*}{\textbf{Data}} & \multicolumn{8}{c}{\textbf{Algorithms' mean run-times}}\\
    	& Ga-NU & Ga-BM & Pso-NU	 & Pso-BM & Pls-NU & Pls-BM & Prs-NU & Prs-BM\\
      	\toprule
      	APS & 243.12 & 543.65 & 335.65 & 1711.50 & 133.88 & 253.88 & 53.17 & 179.40\\
		AQY & 346.30 & 653.78 & 439.62 & 487.62 & 60.69 & 98.26 & 59.20 & 51.09\\
		C2K & 165.80 & 422.08 & 136.93 & 416.10 & 23.18 & 60.14 & 22.67 & 43.79\\
		DIR & 1475.00 & 3516.75 & 828.90 & 1193.25 & 226.18 & 376.28 & 222.38 & 187.98\\
		HPC & 495.55 & 1632.75 & 489.70 & 628.88 & 114.78 & 96.68 & 92.06 & 77.50\\
		OMD & 293.98 & 767.12 & 123.00 & 135.65 & 38.08 & 52.42 & 35.28 & 29.86\\
		HCV & 0.77 & 1.05 & 0.47 & 0.48 & 0.22 & 0.21 & 0.19 & 0.15\\
		CPX & 1.73 & 2.34 & 1.61 & 1.58 & 0.33 & 0.31 & 0.28 & 0.22\\
		BCR & 0.36 & 0.39 & 0.52 & 0.48 & 0.08 & 0.09 & 0.08 & 0.07\\
      	\bottomrule
    	\end{tabular}
    \label{tab:exp1_runtime}
	\end{table}
	
	Table~\ref{tab:exp1_runtime} shows the mean run-times for each algorithm on the numeric-based and bitmap-based search space. Using these values, the results of the Wilcoxon test are shown in Table~\ref{tab:exp1_wilcoxon_run}.
	
	\begin{table}[h!]
  	\centering
    \scriptsize
   	\caption{Wilcoxon test on efficacy of search space}
    	\begin{tabular}{l | l}
    	\textbf{Algorithm} & Results\\
      	\toprule
      	GA-GRAD & test statistic = 0, p-value = 0.0039\\
      	PSO-GRAD & test statistic = 5, p-value = 0.0391\\
      	LS-GRAD & test statistic = 10, p-value = 0.1641\\
      	RS-GRAD & test statistic = 16, p-value = 0.4961\\
      	\bottomrule
    	\end{tabular}
    \label{tab:exp1_wilcoxon_run}
	\end{table}

	\subsubsection{Experiment 2: Comparing Candidate GPs from Search Spaces}
	\label{sec:exp_scatter}
\begin{figure}[h!]
		\centering
		\scriptsize
		\includegraphics[width=\textwidth]{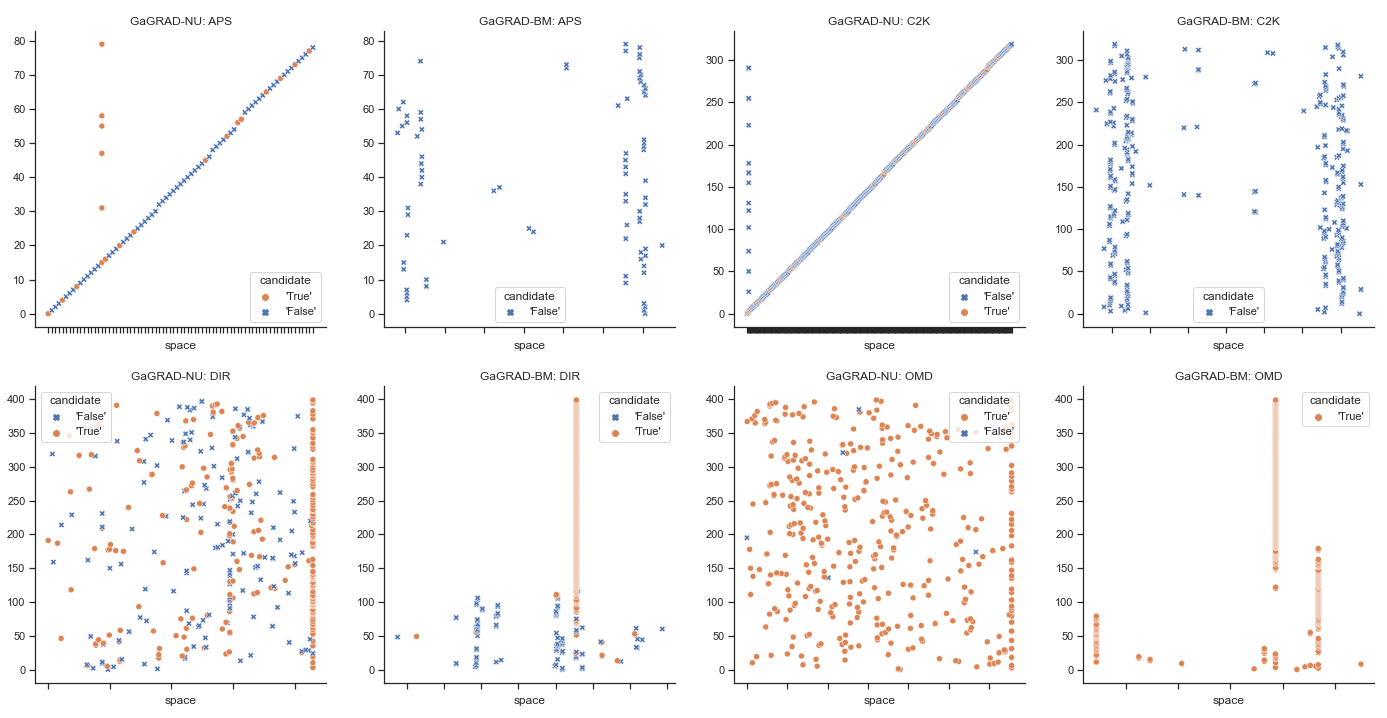}
		\caption{GA-GRAD's candidates from APS, C2K, DIR and OMD data sets using numeric-based and bitmap-based search spaces.}
		\label{fig:exp2_ga1}
	\end{figure}

\begin{figure}[h!]
		\centering
		\scriptsize
		\includegraphics[width=\textwidth]{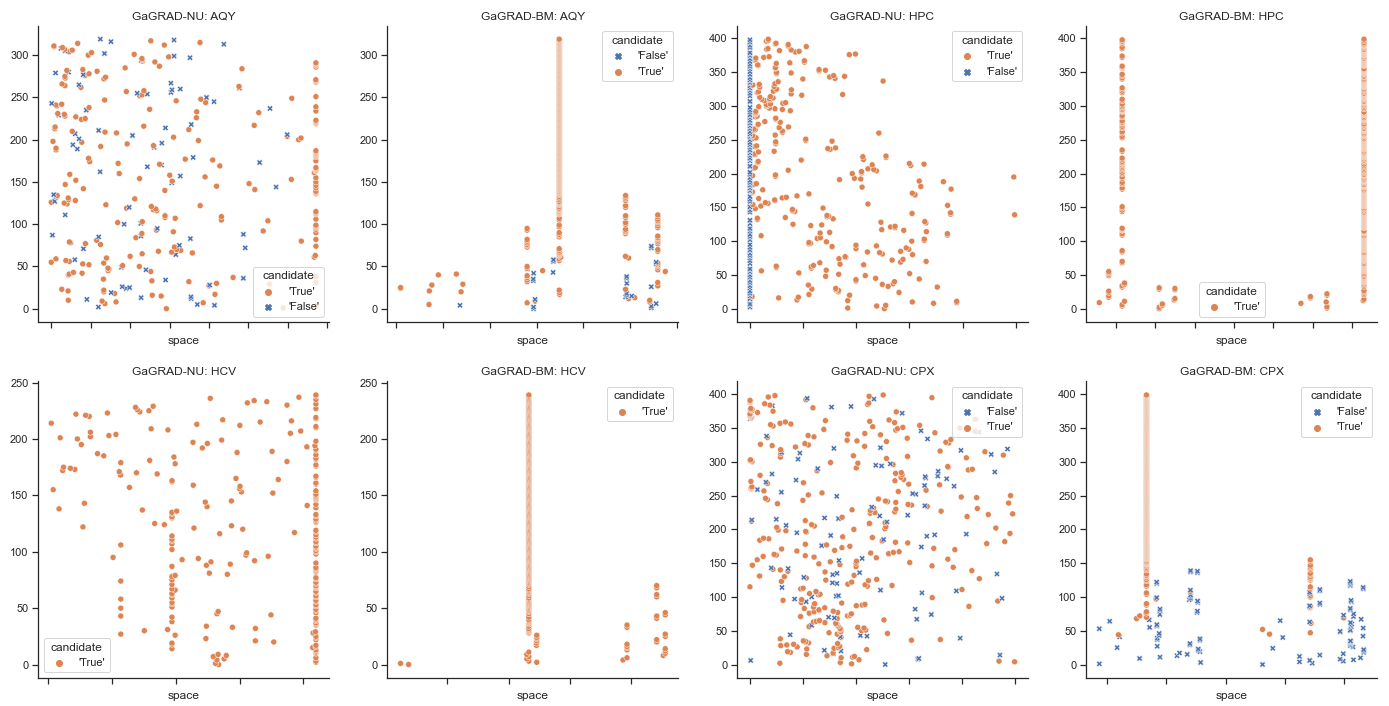}
		\caption{GA-GRAD's candidates from AQY, HPC, HCV and CPX data sets.}
		\label{fig:exp2_ga2}
	\end{figure}

\begin{figure}[h!]
		\centering
		\scriptsize
		\includegraphics[width=\textwidth]{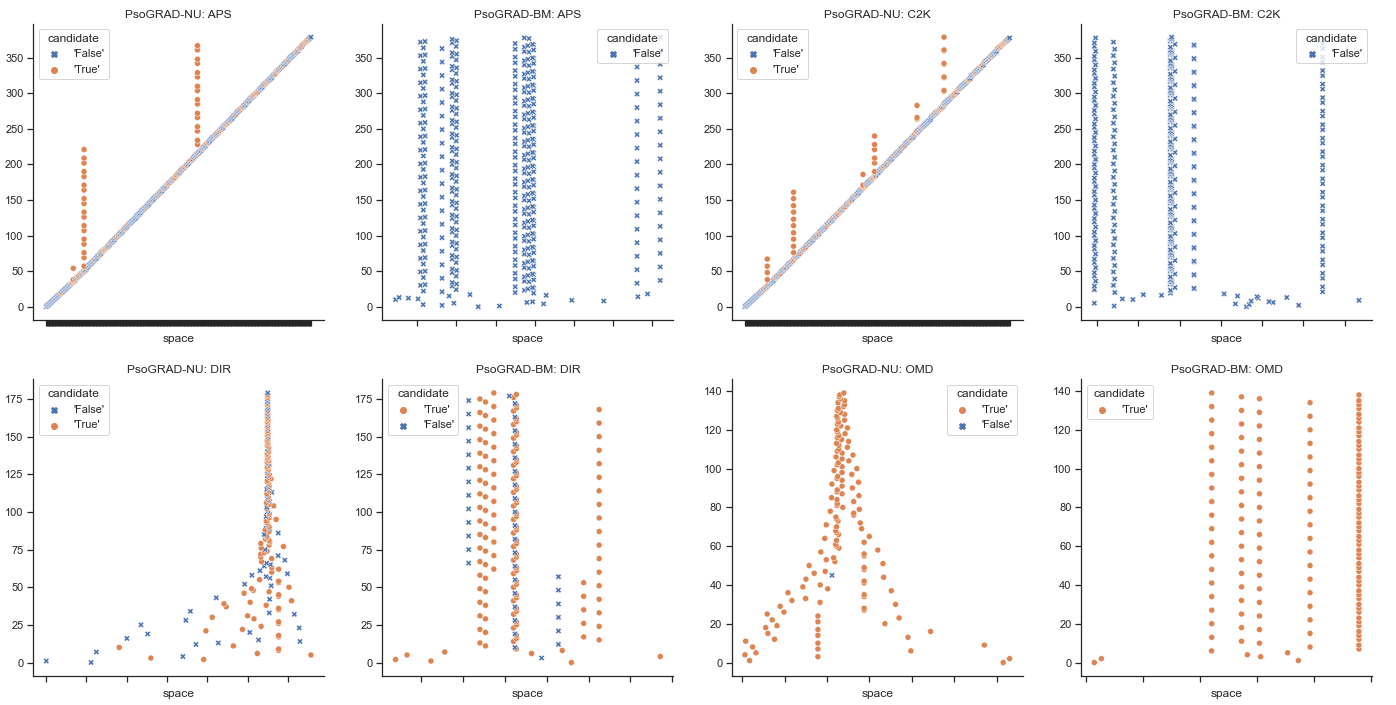}
		\caption{PSO-GRAD's candidates from APS, C2K, DIR and OMD data sets using numeric-based and bitmap-based search spaces.}
		\label{fig:exp2_pso1}
	\end{figure}

\begin{figure}[h!]
		\centering
		\scriptsize
		\includegraphics[width=\textwidth]{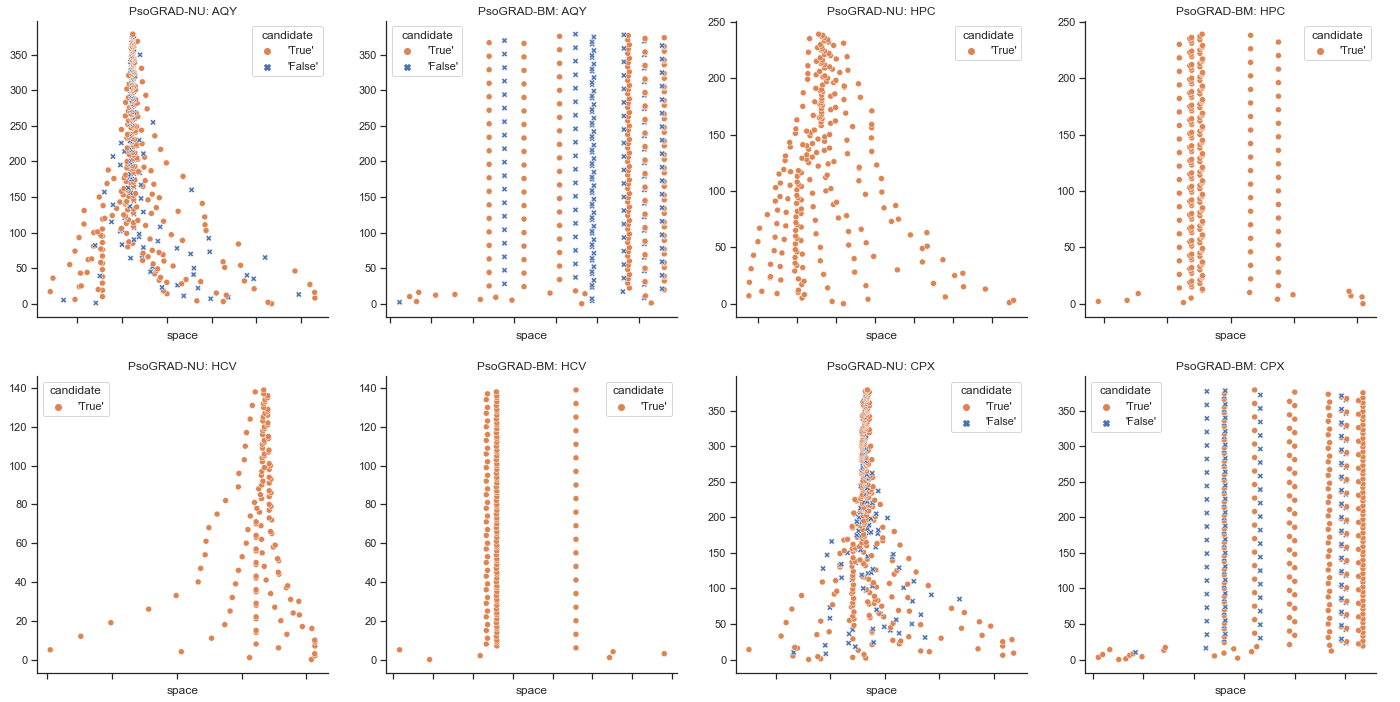}
		\caption{PSO-GRAD's candidates from AQY, HPC, HCV and CPX data sets using numeric-based and bitmap-based search spaces.}
		\label{fig:exp2_pso2}
	\end{figure}

\begin{figure}[h!]
		\centering
		\scriptsize
		\includegraphics[width=\textwidth]{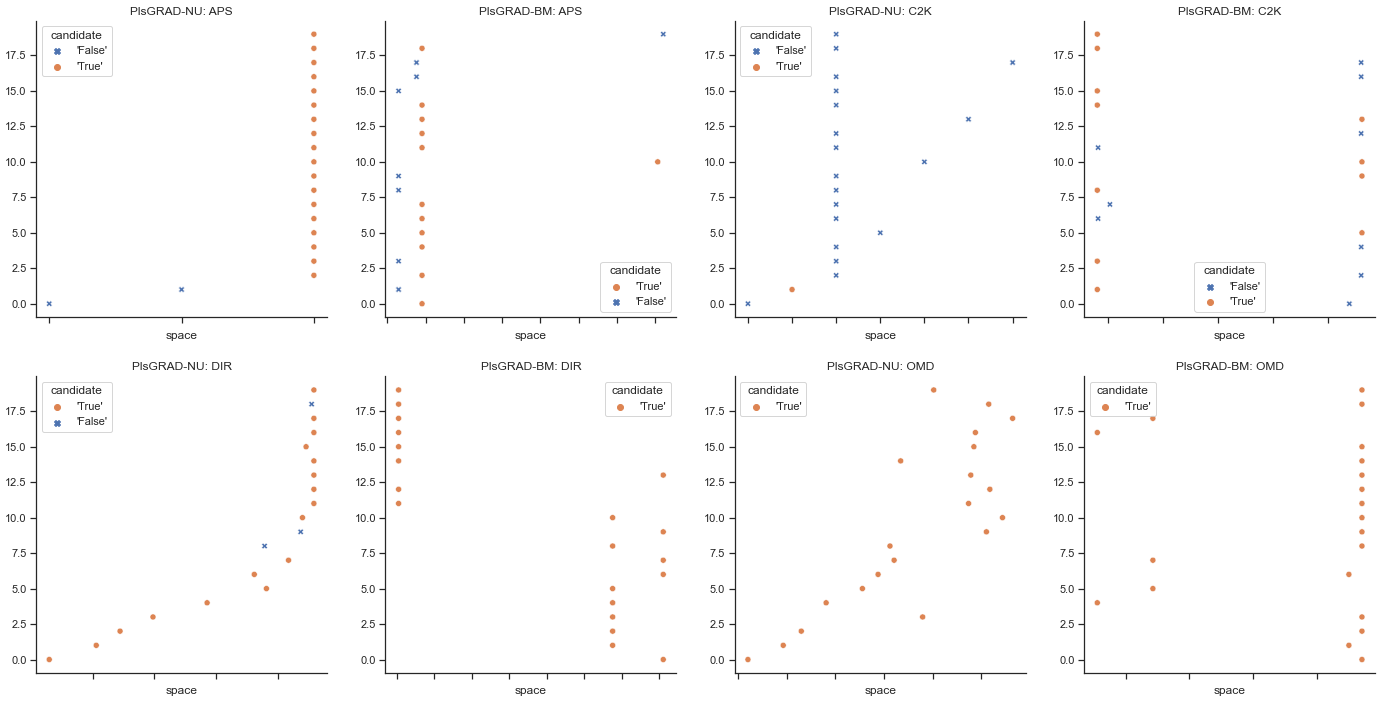}
		\caption{LS-GRAD's candidates from APS, C2K, DIR and OMD data sets using numeric-based and bitmap-based search spaces.}
		\label{fig:exp2_pls1}
	\end{figure}

\begin{figure}[h!]
		\centering
		\scriptsize
		\includegraphics[width=\textwidth]{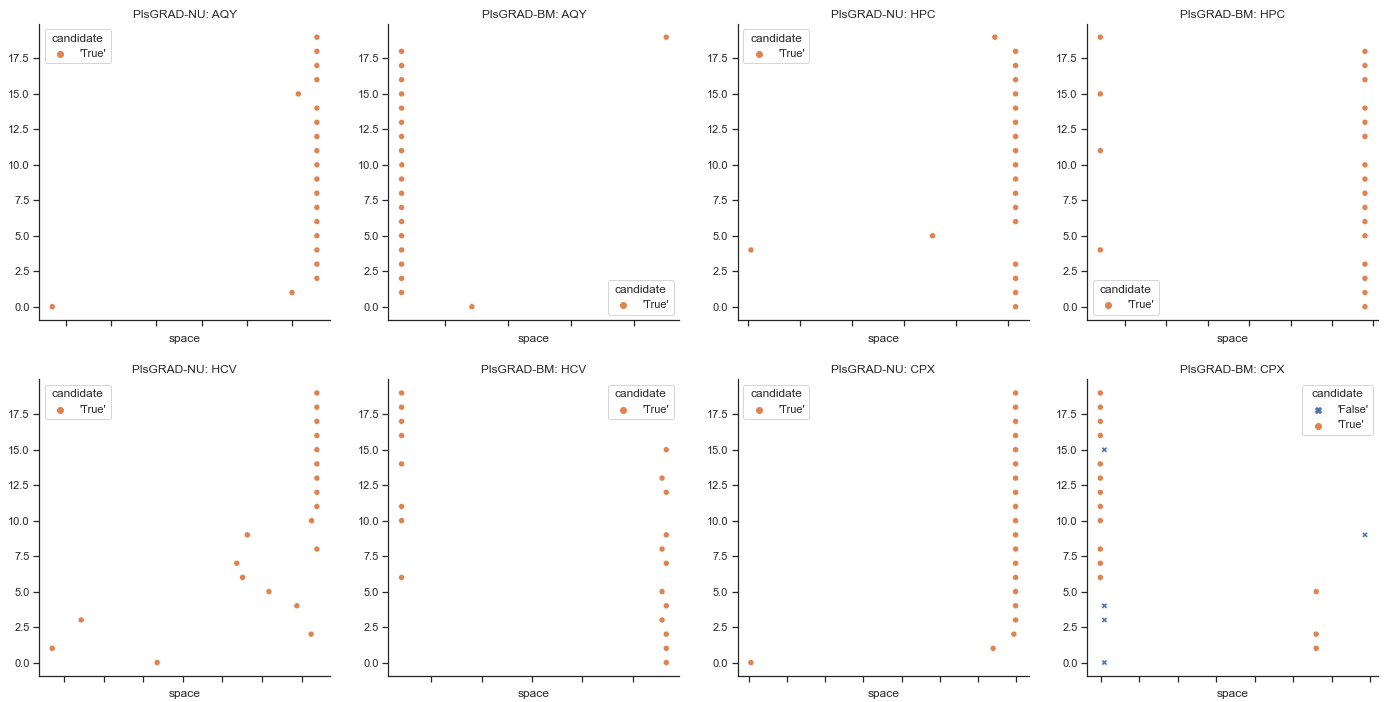}
		\caption{LS-GRAD's candidates from AQY, HPC, HCV and CPX data sets using numeric-based and bitmap-based search spaces.}
		\label{fig:exp2_pls2}
	\end{figure}

\begin{figure}[h!]
		\centering
		\scriptsize
		\includegraphics[width=\textwidth]{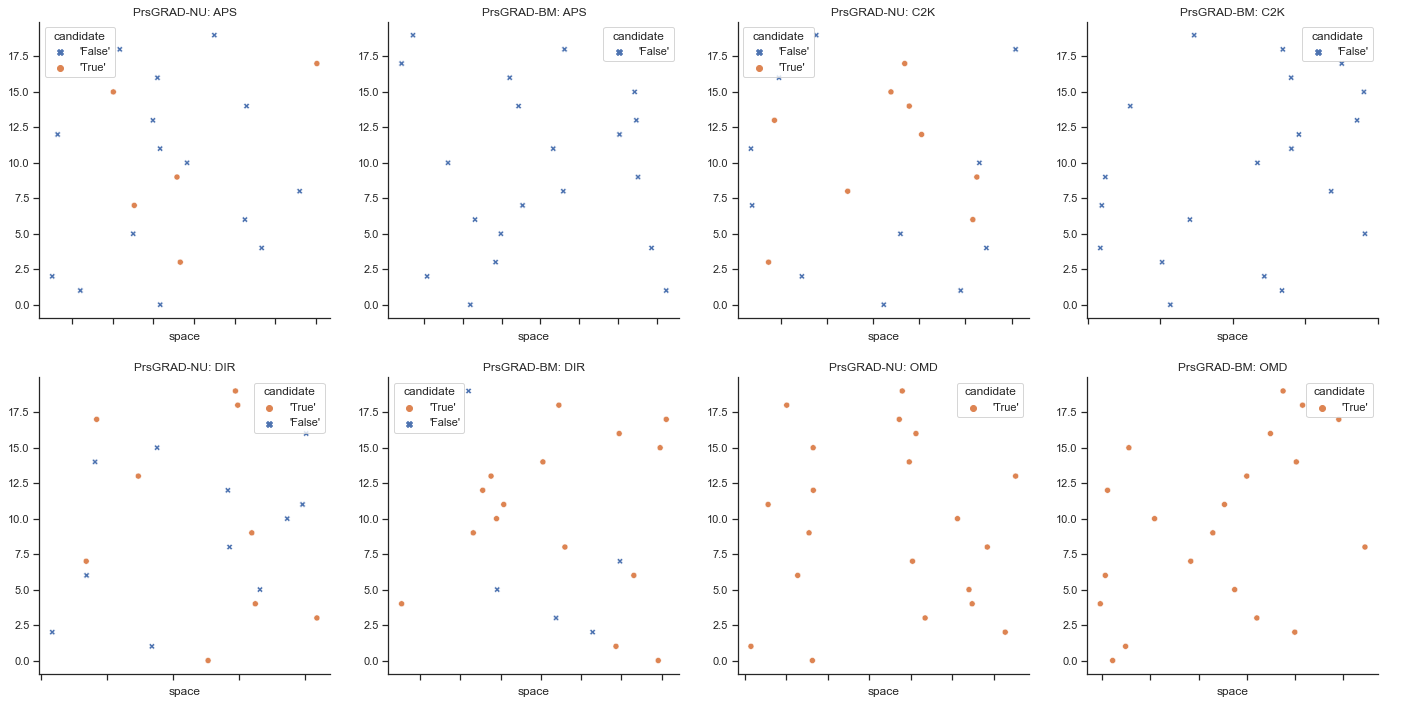}
		\caption{RS-GRAD's candidates from APS, C2K, DIR and OMD data sets using numeric-based and bitmap-based search spaces.}
		\label{fig:exp2_prs1}
	\end{figure}

\begin{figure}[h!]
		\centering
		\scriptsize
		\includegraphics[width=\textwidth]{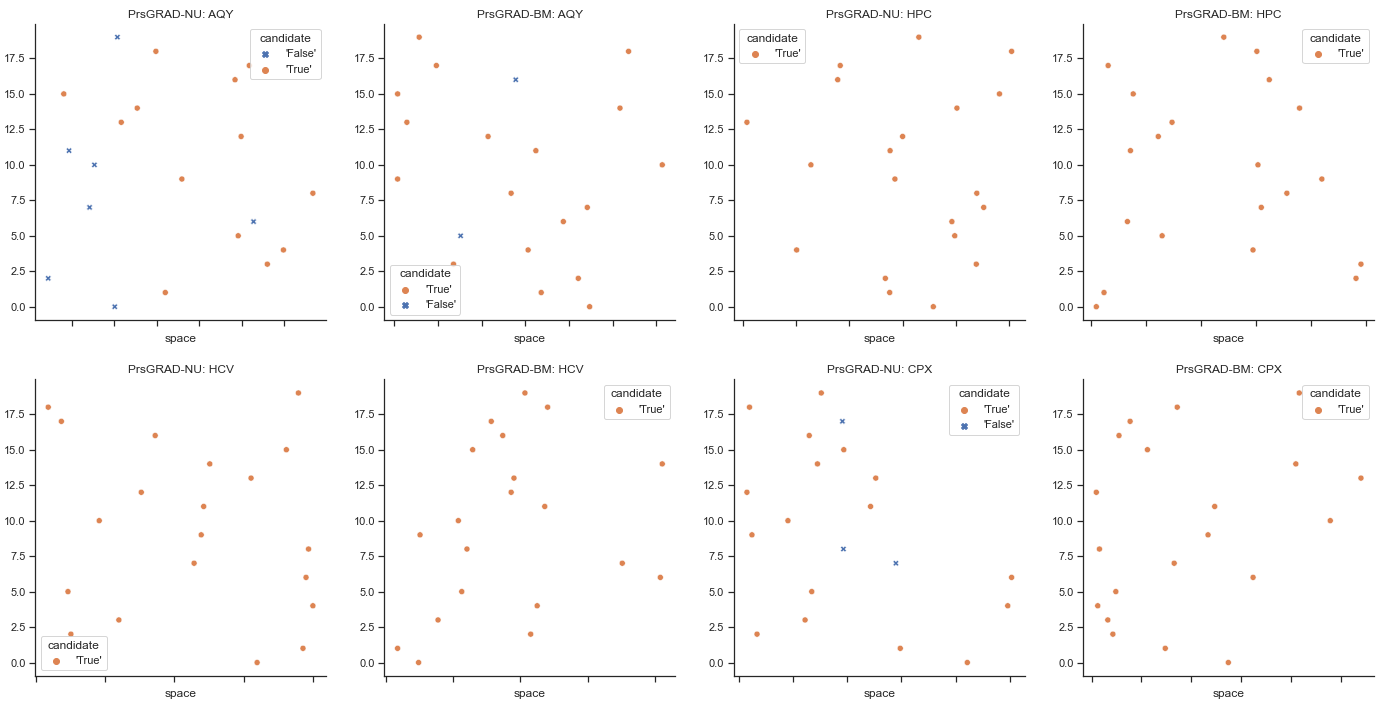}
		\caption{RS-GRAD's candidates from AQY, HPC, HCV and CPX data sets using numeric-based and bitmap-based search spaces.}
		\label{fig:exp2_prs2}
	\end{figure}

	Here, we seek to discover which search space allows our algorithms to yield a better distribution in favor of \textit{valid} GPs. Every candidate that each algorithm generates, is evaluated to check if it is valid or invalid. Figures~\ref{fig:exp2_ga1}, \ref{fig:exp2_ga2}, \ref{fig:exp2_pso1}, \ref{fig:exp2_pso2}, \ref{fig:exp2_pls1}, \ref{fig:exp2_pls2}, \ref{fig:exp2_prs1}, \ref{fig:exp2_prs2} are scatter plots of position against fitness/validity of candidates, where valid GPs are labelled as \textit{`True'} and invalid GPs are labelled as \textit{`False'}.
	
	In Figures~\ref{fig:exp2_ga1} and \ref{fig:exp2_ga2}, we observe that the distribution of GP candidates is almost uniformly spread across the entire numeric-based search space; while, it is sparsely spread across the bitmap-based search space. APS and C2K data sets have large number of attributes in comparison to the other data sets; this explains why the distribution of candidates diagonally cuts across the numeric search space. GA-GRAD algorithm does not find any valid GP using the bitmap-based search space from APS and C2K data sets.

	\clearpage
	
	Similarly in the case of PSO-GRAD algorithm, Figure~\ref{fig:exp2_pso1} shows that PSO-GRAD algorithm does not find any valid GP using the bitmap-based search space from APS and C2K data sets.
	
	Again in Figures~\ref{fig:exp2_pso1} and \ref{fig:exp2_pso2}, we observe that the distribution of candidates are directed towards a single point in the case of the numeric-based search space.

	It is interesting to observe in Figures~\ref{fig:exp2_pls1} and \ref{fig:exp2_pls2}, that there is no significant difference between the distributions of GP candidates in both the numeric-based and the bitmap-based search spaces. This phenomenon confirms the results of the Wilcoxon test presented in Section~\ref{sec:exp_wilcoxon} concerning the LS-GRAD algorithm.

	In the case of RS-GRAD algorithm, we observe in Figures~\ref{fig:exp2_prs1} and \ref{fig:exp2_prs2}, that there is no significant difference between the distributions of GP candidates in both the numeric-based and the bitmap-based search spaces. Again this observation confirms the results of the Wilcoxon test presented in Section~\ref{sec:exp_wilcoxon}.
	
	%\clearpage
	%\newpage
	
	\subsubsection{Experiment 3: Comparative Analysis}
	\label{sec:exp_compare}
	In this experiment, we compare the performance of our algorithms in terms of run-time duration, memory consumption and number of extracted GPs against: GRAANK initially proposed by \cite{Laurent2009} and ACO-GRAANK proposed by \cite{Owuor2021jul}.
	
	\textcolor{brown}{We clarify the following: (1) There exists other GP mining algorithms that rely on a tree-based search for candidates; such as, GRITE proposed by \cite{Laurent2010} and ParaMiner proposed by \cite{Negrevergne2014}. The gradualness semantic of such algorithms (which is focused on an ordered precedence graph of objects) is different from those under investigation in this study (which extend GRAANK), in that they rely on a breath-first search and use a gradualness semantic focused on concordant object couples. The proposed search spaces are best suited for GRAANK-based algorithms; therefore, it is almost unfair to compare their computational performance to GRITE-based algorithms. (2) We use AcoGRAD to denote ACO-GRAANK. (3) GRAANK algorithm yields \textit{`Memory Error'} when executed on APS and DIR data sets.}
		
	\begin{figure}[h!]
		\centering
		\scriptsize
		% First 
		\subfloat[]{%
		\includegraphics[scale=0.24]{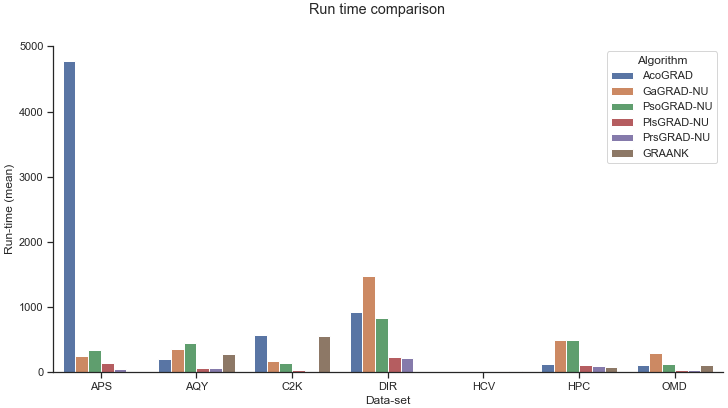}}
		% Second 
		\subfloat[]{%
		\includegraphics[scale=0.24]{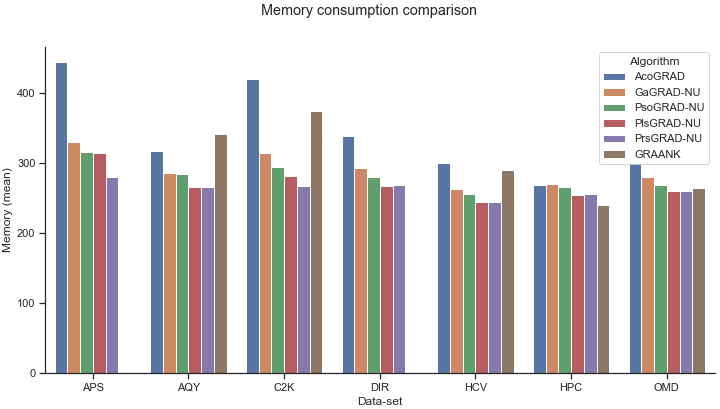}}
		\caption{Comparison of algorithms' run-times and memory consumptions.}
		\label{fig:exp3_run_mem}
	\end{figure}
	
	In Figure~\ref{fig:exp3_run_mem} (a), we observe that AcoGRAD has the longest run-time when applied on data sets with large number of attributes (i.e., APS and C2K). However, AcoGRAD and GRAANK algorithms have the shorter run-times than PSO-GRAD and GA-GRAD algorithms when applied on data sets with few attributes (i.e., HPC, OMD, AQY). LS-GRAD and RS-GRAD algorithms have the shortest run-times across all the data sets. In terms of memory consumption (as shown in Figure~\ref{fig:exp3_run_mem} (b)), with the exception of AcoGRAD and GRAANK, all the other algorithms consume almost the same amount of memory. However, GRAANK's memory consumption grows exponentially when applied on APS and DIR data sets; thus, yielding a \textit{`Memory Error'}. 
	
\begin{figure}[h!]
		\centering
		\scriptsize
		% First 
		\subfloat[]{%
		\includegraphics[scale=0.24]{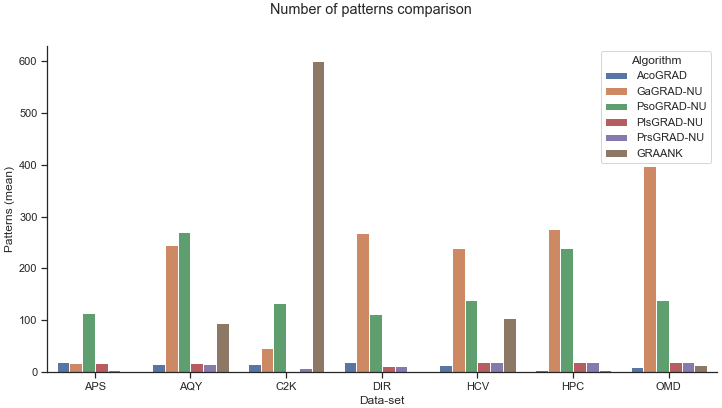}}
		% Second 
		\subfloat[]{%
		\includegraphics[scale=0.24]{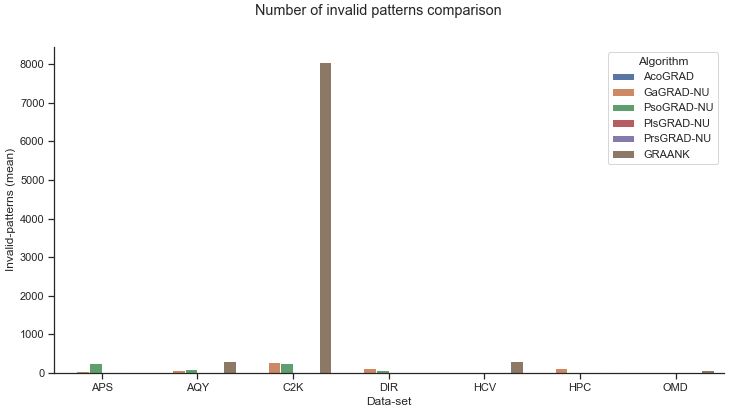}}
		\caption{Comparison of number of GPs extracted by each algorithm.}
		\label{fig:exp3_patterns}
	\end{figure}

	In Figure~\ref{fig:exp3_patterns} (a), we observe that GRAANK extracts the most number of valid GPs in C2K data set. However, in all the other data sets GA-GRAD and PSO-GRAD algorithms extract the most number of valid GPs. Recall that GRAANK algorithm yields \textit{`Memory Error'} when applied on APS and DIR data set; for this reason, the number of valid and invalid GPs could not be recorded.
	
	In Figure~\ref{fig:exp3_patterns} (b), we observe that GRAANK algorithm generates the most number of invalid GPs especially in C2K, AQY and HCV data sets. It is interesting to observe that our meta-heuristic algorithms generally generate very few invalid GPs. However, we also observe that GA-GRAD and PSO-GRAD algorithms generate slightly more number of invalid GPs than RS-GRAD and LS-GRAD algorithms.
	
	%\clearpage	
	\subsection{Discussion of Results}
	\label{sec:discussion}
	
	\subsubsection{Search Space}
	\label{sec:discussion_space}
	The Wilcoxon test results presented in Table~\ref{tab:exp1_wilcoxon_run} show low p-value (less than 0.05) for GA-GRAD and PSO-GRAD algorithms. This implies that there is significant differences in the run-time durations when these 2 algorithms are using the bitmap-based search space versus when they are using the numeric-based search space. From Table~\ref{tab:exp1_runtime}, we observe that al 4 algorithms have faster run-times in most data sets when using the numeric-based search space. This makes the numeric-based search space more effective than its counterpart especially in the case of PSO-GRAD and GA-GRAD algorithms.
	
	Again in Section~\ref{sec:exp_scatter}, we observe that in the scatter plots of GA-GRAD and PSO-GRAD algorithms, GP candidates are more uniformly spread in the numeric-based search space than in the bitmap-based search space. This implies that the numeric-based search space allows these algorithms to explore more areas of the search space. Furthermore, these 2 algorithms do not find any valid GP from data sets with large numbers of attributes using the bitmap-based search space (i.e., APS and C2K). This may confirm the inefficacy of the bitmap-based search space especially when used on data sets with large numbers of data sets.
	
	However, LS-GRAD and RS-GRAD algorithms seem to have almost similar distribution of GP candidates and little differences in run-time durations in both search spaces. In order to break the tie in favor of the numeric-based search spaces, we focus on the APS and C2K data sets. We observe in Figures~\ref{fig:exp2_pls1} and \ref{fig:exp2_prs1} that both algorithms do not find any valid GP in the APS data set. Although LS-GRAD algorithm finds valid GPs from C2K data set using both search spaces, it does so at the expense of a longer run-time in the bitmap-based search space (see Table~\ref{tab:exp1_runtime}). In fact, the run-time is almost 3 times as long in comparison to the numeric-based search space.
		
	\subsubsection{Execution Run-time}
	\label{sec:discussion_runtime}
	We observe that generally PSO-GRAD and GA-GRAD algorithms have longer run times than LS-GRAD and RS-GRAD algorithms. Specifically, PSO-GRAD followed by GA-GRAD algorithms have the longest run times in all the search spaces. This may be due to the fact that for every search iteration, PSO-GRAD and GA-GRAD algorithms evaluate more than one candidate GP. 
	
	Recall (from Section~\ref{sec:ga_pso_complexity} and Section~\ref{sec:ls_rs_complexity}) that the task of evaluating GP candidates is the most computationally intensive task. Therefore, the more GP candidates that are evaluated by the \textit{objective function} in every search iteration, the higher the computational complexity of the algorithm.
	
	\subsubsection{Extracted Patterns}
	\label{sec:discussion_patterns}
	We observe that generally GA-GRAD, PSO-GRAD and RS-GRAD algorithms extract more GPs than LS-GRAD algorithm. This phenomenon may be attributed to the fact PSO-GRAD and GA-GRAD algorithms generally evaluate higher number of candidates per every iteration in comparison to LS-GRAD and RS-GRAD algorithm, which in turn improves their chances of finding more valid GPs. \textcolor{blue}{However, we mention that our proposed metaheuristic algorithms are not guaranteed to find all frequent GPs in each data set. In reality, their searches only focus and converge to candidates in the search space that extract the GPs with the highest support (also referred to as best GPs), which is a subset of the frequent GPs.}
	
	\textcolor{blue}{In contrast, the classical GRAANK (initially proposed by \cite{Laurent2009}) is a complete GP mining algorithm; in the sense that, it tests all possible candidates in the search space and extracts all the frequent GPs.} In fact, our experiments reveal that GRAANK algorithm extracts the most number of valid GPs (especially in C2K data set); however, this comes with the downside of a higher run-time, an exponential growth of memory consumption (especially for APS and DIR data sets), and a generation of the huge numbers of invalid GPs. 
	
	\textcolor{blue}{Given these points, we recommend GRAANK algorithm as the suitable algorithm for extracting GPs from small data sets and, we recommend our proposed metaheuristic algorithms (i.e., GA-GRAD, PSO-GRAD, LS-GRAD and RS-GRAD)  as the suitable algorithms for extracting GPs from large data sets (especially those with huge numbers of attributes).}

	\section{Conclusion and Future Works}
	\label{sec:conclusion}
	In this research study, we examine the problem of finding valid gradual patterns from a huge search space without having to test every candidate. We identify stochastic, evolutionary and swarm based optimization techniques as one of the efficient techniques for solving this problem. We define a lean search space based on numeric encoding that can easily be explored by these algorithms to find valid GPs.
		
	In our experimental study, we compare the performances of LS-GRAD, RS-GRAD algorithms, GA-GRAD and PSO-GRAD algorithms in exploring gradual pattern search spaces based on bitmap encoding verses those based on numeric encoding. The experiment results reveal that all the algorithms find best gradual patterns at shorter run-times from search spaces built on numeric encoding. This may imply that the numeric encoding builds search spaces that are leaner and easier to explore by meta-heuristic algorithms than the bitmap encoding.
		
	In future, an even more effective encoding for gradual pattern candidates may be envisioned that allows gradual patterns to be extracted using machine learning techniques. It may also be interesting to explore the possibility of further reducing the search space of candidates so that algorithms generate even fewer invalid gradual patterns.

	\section*{Acknowledgment}
	This work has been realized with the support of the High Performance Computing Platform: \textbf{MESO@LR}\footnote{\url{https://meso-lr.umontpellier.fr}}, financed by the Occitanie / Pyrénées-Méditerranée Region, Montpellier Mediterranean Metropole and Montpellier University.

	\bibliography{bibliography}

\begin{thebibliography}{10}
\expandafter\ifx\csname url\endcsname\relax
  \def\url#1{\texttt{#1}}\fi
\expandafter\ifx\csname urlprefix\endcsname\relax\def\urlprefix{URL }\fi
\expandafter\ifx\csname href\endcsname\relax
  \def\href#1#2{#2} \def\path#1{#1}\fi

\bibitem{Wahab2015}
M.~N. Ab~Wahab, S.~Nefti-Meziani, A.~Atyabi, A comprehensive review of swarm
  optimization algorithms, PLOS ONE 10~(5) (2015) 1--36.
\newblock \href {https://doi.org/10.1371/journal.pone.0122827}
  {\path{doi:10.1371/journal.pone.0122827}}.

\bibitem{Mohamed2018}
M.~Abdel-Basset, L.~Abdel-Fatah, A.~K. Sangaiah, Chapter 10 - metaheuristic
  algorithms: A comprehensive review, in: A.~K. Sangaiah, M.~Sheng, Z.~Zhang
  (Eds.), Computational Intelligence for Multimedia Big Data on the Cloud with
  Engineering Applications, Intelligent Data-Centric Systems, Academic Press,
  2018, pp. 185--231.
\newblock \href
  {https://doi.org/https://doi.org/10.1016/B978-0-12-813314-9.00010-4}
  {\path{doi:https://doi.org/10.1016/B978-0-12-813314-9.00010-4}}.

\bibitem{Chen2018}
C.~Fan, Y.~Sun, K.~Shan, F.~Xiao, J.~Wang, Discovering gradual patterns in
  building operations for improving building energy efficiency, Applied Energy
  224 (2018) 116--123.
\newblock \href
  {https://doi.org/https://doi.org/10.1016/j.apenergy.2018.04.118}
  {\path{doi:https://doi.org/10.1016/j.apenergy.2018.04.118}}.

\bibitem{Di-Jorio2008}
L.~Di~Jorio, A.~Laurent, M.~Teisseire, Fast extraction of gradual association
  rules: A heuristic based method, in: Proceedings of the 5th International
  Conference on Soft Computing as Transdisciplinary Science and Technology,
  CSTST '08, Association for Computing Machinery, New York, NY, USA, 2008, p.
  205–210.
\newblock \href {https://doi.org/10.1145/1456223.1456268}
  {\path{doi:10.1145/1456223.1456268}}.

\bibitem{Owuor2021jul}
D.~O. Owuor, T.~Runkler, A.~Laurent, J.~O. Orero, E.~O. Menya, Ant colony
  optimization for mining gradual patterns, International Journal of Machine
  Learning and Cybernetics (2021) 1--21\href
  {https://doi.org/10.1007/s13042-021-01390-w}
  {\path{doi:10.1007/s13042-021-01390-w}}.

\bibitem{Obitko1999}
M.~Obitko, P.~Slav{\'\i}k, Visualization of genetic algorithms in a learning
  environment, in: Spring Conference on Computer Graphics, SCCG, Vol.~99, 1999,
  pp. 101--106.

\bibitem{Patel2021}
V.~P. Patel, M.~K. Rawat, A.~S. Patel, Analysis of search space in the domain
  of swarm intelligence, in: M.~Prateek, T.~P. Singh, T.~Choudhury, H.~M.
  Pandey, N.~Gia~Nhu (Eds.), Proceedings of International Conference on Machine
  Intelligence and Data Science Applications, Springer Singapore, Singapore,
  2021, pp. 99--109.

\bibitem{Berzal2007}
F.~Berzal, J.~C. Cubero, D.~Sanchez, M.~A. Vila, J.~M. Serrano, An alternative
  approach to discover gradual dependencies, International Journal of
  Uncertainty, Fuzziness and Knowledge-Based Systems 15~(05) (2007) 559--570.
\newblock \href {https://doi.org/10.1142/S021848850700487X}
  {\path{doi:10.1142/S021848850700487X}}.

\bibitem{Laurent2009}
A.~Laurent, M.-J. Lesot, M.~Rifqi, Graank: Exploiting rank correlations for
  extracting gradual itemsets, in: Proceedings of the 8th International
  Conference on Flexible Query Answering Systems, FQAS '09, Springer-Verlag,
  Berlin, Heidelberg, 2009, pp. 382--393.
\newblock \href {https://doi.org/10.1007/978-3-642-04957-6_33}
  {\path{doi:10.1007/978-3-642-04957-6_33}}.

\bibitem{Owuor2021aug}
D.~O. Owuor, A.~Laurent, Efficiently mining large gradual patterns using
  chunked storage layout, in: L.~Bellatreche, M.~Dumas, P.~Karras,
  R.~Matulevi{\v{c}}ius (Eds.), Advances in Databases and Information Systems,
  Springer International Publishing, Cham, 2021, pp. 30--42.
\newblock \href {https://doi.org/10.1007/978-3-030-82472-3_4}
  {\path{doi:10.1007/978-3-030-82472-3_4}}.

\bibitem{Owuor2019}
D.~{Owuor}, A.~{Laurent}, J.~{Orero}, Mining fuzzy-temporal gradual patterns,
  in: 2019 IEEE International Conference on Fuzzy Systems (FUZZ-IEEE), IEEE,
  New York, NY, USA, 2019, pp. 1--6.
\newblock \href {https://doi.org/10.1109/FUZZ-IEEE.2019.8858883}
  {\path{doi:10.1109/FUZZ-IEEE.2019.8858883}}.

\bibitem{Owuor2021oct}
D.~O. Owuor, A.~Laurent, J.~O. Orero, Mining fuzzy temporal gradual emerging
  patterns, International Journal of Uncertainty, Fuzziness and Knowledge-Based
  Systems (2021).
\newblock \href {https://doi.org/10.1142/S0218488521500288}
  {\path{doi:10.1142/S0218488521500288}}.

\bibitem{Di-Jorio2009}
L.~Di-Jorio, A.~Laurent, M.~Teisseire, Mining frequent gradual itemsets from
  large databases, in: Advances in Intelligent Data Analysis VIII,
  Springer-Verlag, Berlin, Heidelberg, 2009, pp. 297--308.
\newblock \href {https://doi.org/10.1007/978-3-642-03915-7_26}
  {\path{doi:10.1007/978-3-642-03915-7_26}}.

\bibitem{Clementin2021}
T.~D. Clémentin, T.~F.~L. Cabrel, K.~E. Belise, A novel algorithm for
  extracting frequent gradual patterns, Machine Learning with Applications 5
  (2021) 100068.
\newblock \href {https://doi.org/10.1016/j.mlwa.2021.100068}
  {\path{doi:10.1016/j.mlwa.2021.100068}}.

\bibitem{Negrevergne2014}
B.~Negrevergne, A.~Termier, M.-C. Rousset, J.-F. M{\'e}haut, Paraminer: a
  generic pattern mining algorithm for multi-core architectures, Data Mining
  and Knowledge Discovery 28~(3) (2014) 593--633.
\newblock \href {https://doi.org/10.1007/s10618-013-0313-2}
  {\path{doi:10.1007/s10618-013-0313-2}}.

\bibitem{Owuor2021jan}
D.~Owuor, A.~Laurent, J.~Orero, O.~Lobry, Gradual pattern mining tool on cloud,
  Extraction et Gestion des Connaissances: Actes EGC'2021 (2021).

\bibitem{Solis1981}
F.~J. Solis, R.~J.-B. Wets, Minimization by random search techniques,
  Mathematics of Operations Research 6~(1) (1981) 19--30.
\newblock \href {https://doi.org/10.1287/moor.6.1.19}
  {\path{doi:10.1287/moor.6.1.19}}.

\bibitem{Zabinsky2003}
Z.~B. Zabinsky, Pure Random Search and Pure Adaptive Search, Springer US,
  Boston, MA, 2003, pp. 25--54.
\newblock \href {https://doi.org/10.1007/978-1-4419-9182-9_2}
  {\path{doi:10.1007/978-1-4419-9182-9_2}}.

\bibitem{Zabinsky2011}
Z.~Zabinsky, Random Search Algorithms, American Cancer Society, 2011.
\newblock \href {https://doi.org/10.1002/9780470400531.eorms0704}
  {\path{doi:10.1002/9780470400531.eorms0704}}.

\bibitem{Franti2000}
P.~Fr{\"a}nti, J.~Kivij{\"a}rvi, Randomised local search algorithm for the
  clustering problem, Pattern Analysis \& Applications 3~(4) (2000) 358--369.
\newblock \href {https://doi.org/10.1007/s100440070007}
  {\path{doi:10.1007/s100440070007}}.

\bibitem{Hoos2004}
H.~H. Hoos, T.~St{\"u}tzle, Stochastic local search: Foundations and
  applications, Elsevier, 2004.

\bibitem{Ishibuchi1996}
H.~Ishibuchi, T.~Murata, Multi-objective genetic local search algorithm, in:
  Proceedings of IEEE International Conference on Evolutionary Computation,
  1996, pp. 119--124.
\newblock \href {https://doi.org/10.1109/ICEC.1996.542345}
  {\path{doi:10.1109/ICEC.1996.542345}}.

\bibitem{Johnson1988}
D.~S. Johnson, C.~H. Papadimitriou, M.~Yannakakis, How easy is local search?,
  Journal of Computer and System Sciences 37~(1) (1988) 79--100.
\newblock \href {https://doi.org/10.1016/0022-0000(88)90046-3}
  {\path{doi:10.1016/0022-0000(88)90046-3}}.

\bibitem{Hossain2014}
M.~Hossain, T.~Tasnim, S.~Shatabda, D.~M. Farid, Stochastic local search for
  pattern set mining, in: The 8th International Conference on Software,
  Knowledge, Information Management and Applications (SKIMA 2014), 2014, pp.
  1--6.
\newblock \href {https://doi.org/10.1109/SKIMA.2014.7083547}
  {\path{doi:10.1109/SKIMA.2014.7083547}}.

\bibitem{Holland1975}
J.~Holland, Adaptation in natural and artificial systems, univ. of mich. press,
  Ann Arbor (1975).

\bibitem{Koza1992}
J.~R. Koza, J.~R. Koza, Genetic programming: on the programming of computers by
  means of natural selection, Vol.~1, MIT press, 1992.

\bibitem{Mirjalili2019}
S.~Mirjalili, Genetic Algorithm, Springer International Publishing, Cham, 2019,
  pp. 43--55.
\newblock \href {https://doi.org/10.1007/978-3-319-93025-1_4}
  {\path{doi:10.1007/978-3-319-93025-1_4}}.

\bibitem{Kabir2015}
M.~M.~J. Kabir, S.~Xu, B.~H. Kang, Z.~Zhao, Comparative analysis of genetic
  based approach and apriori algorithm for mining maximal frequent item sets,
  in: 2015 IEEE Congress on Evolutionary Computation (CEC), 2015, pp. 39--45.
\newblock \href {https://doi.org/10.1109/CEC.2015.7256872}
  {\path{doi:10.1109/CEC.2015.7256872}}.

\bibitem{Saravanan2014}
M.~Saravanan, V.~L. Jyothi, A novel approach for sequential pattern mining by
  using genetic algorithm, in: 2014 International Conference on Control,
  Instrumentation, Communication and Computational Technologies (ICCICCT),
  2014, pp. 284--288.
\newblock \href {https://doi.org/10.1109/ICCICCT.2014.6992971}
  {\path{doi:10.1109/ICCICCT.2014.6992971}}.

\bibitem{Kennedy1995}
J.~Kennedy, R.~Eberhart, Particle swarm optimization, in: Proceedings of
  ICNN'95 - International Conference on Neural Networks, Vol.~4, 1995, pp.
  1942--1948 vol.4.
\newblock \href {https://doi.org/10.1109/ICNN.1995.488968}
  {\path{doi:10.1109/ICNN.1995.488968}}.

\bibitem{Rajamohana2018}
S.~Rajamohana, K.~Umamaheswari, Hybrid approach of improved binary particle
  swarm optimization and shuffled frog leaping for feature selection, Computers
  \& Electrical Engineering 67 (2018) 497--508.
\newblock \href
  {https://doi.org/https://doi.org/10.1016/j.compeleceng.2018.02.015}
  {\path{doi:https://doi.org/10.1016/j.compeleceng.2018.02.015}}.

\bibitem{Shruti2011}
S.~Mishra, D.~Mishra, S.~K. Satapathy, Particle swarm optimization based fuzzy
  frequent pattern mining from gene expression data, in: 2011 2nd International
  Conference on Computer and Communication Technology (ICCCT-2011), 2011, pp.
  15--20.
\newblock \href {https://doi.org/10.1109/ICCCT.2011.6075204}
  {\path{doi:10.1109/ICCCT.2011.6075204}}.

\bibitem{Shruti2012}
M.~Shruti, M.~Debahuti, S.~Sandeep, Ku., Fuzzy frequent pattern mining from
  gene expression data using dynamic multi-swarm particle swarm optimization,
  Procedia Technology 4 (2012) 797--801, 2nd International Conference on
  Computer, Communication, Control and Information Technology( C3IT-2012).
\newblock \href {https://doi.org/https://doi.org/10.1016/j.protcy.2012.05.130}
  {\path{doi:https://doi.org/10.1016/j.protcy.2012.05.130}}.

\bibitem{Dorigo1996}
M.~{Dorigo}, V.~{Maniezzo}, A.~{Colorni}, Ant system: optimization by a colony
  of cooperating agents, IEEE Transactions on Systems, Man, and Cybernetics,
  Part B (Cybernetics) 26~(1) (1996) 29--41.
\newblock \href {https://doi.org/10.1109/3477.484436}
  {\path{doi:10.1109/3477.484436}}.

\bibitem{Dorigo2019}
M.~Dorigo, T.~St{\"u}tzle, Ant Colony Optimization: Overview and Recent
  Advances, Springer International Publishing, Cham, 2019, pp. 311--351.
\newblock \href {https://doi.org/10.1007/978-3-319-91086-4_10}
  {\path{doi:10.1007/978-3-319-91086-4_10}}.

\bibitem{Runkler2005}
T.~A. Runkler, Ant colony optimization of clustering models, International
  Journal of Intelligent Systems 20~(12) (2005) 1233--1251.
\newblock \href {https://doi.org/10.1002/int.20111}
  {\path{doi:10.1002/int.20111}}.

\bibitem{Stutzle2000}
T.~St{\"u}tzle, H.~H. Hoos, Max--min ant system, Future generation computer
  systems 16~(8) (2000) 889--914.
\newblock \href {https://doi.org/https://doi.org/10.1016/S0167-739X(00)00043-1}
  {\path{doi:https://doi.org/10.1016/S0167-739X(00)00043-1}}.

\bibitem{Harris2010}
D.~Harris, S.~Harris, Digital design and computer architecture, Morgan
  Kaufmann, 2010.

\bibitem{Regan2018}
G.~O'Regan, Binary Number System, Springer International Publishing, Cham,
  2018, pp. 53--56.
\newblock \href {https://doi.org/10.1007/978-3-030-02619-6_12}
  {\path{doi:10.1007/978-3-030-02619-6_12}}.

\bibitem{Nolle2006}
L.~Nolle, On a hill-climbing algorithm with adaptive step size: Towards a
  control parameter-less black-box optimisation algorithm, in: B.~Reusch (Ed.),
  Computational Intelligence, Theory and Applications, Springer Berlin
  Heidelberg, Berlin, Heidelberg, 2006, pp. 587--595.
\newblock \href {https://doi.org/10.1007/3-540-34783-6_56}
  {\path{doi:10.1007/3-540-34783-6_56}}.

\bibitem{Cormen2009}
T.~H. Cormen, C.~E. Leiserson, R.~L. Rivest, C.~Stein, Introduction to
  algorithms, MIT press, 2009.

\bibitem{Vaz2017}
R.~{Vaz}, V.~{Shah}, A.~{Sawhney}, R.~{Deolekar}, Automated big-o analysis of
  algorithms, in: 2017 International Conference on Nascent Technologies in
  Engineering (ICNTE), 2017, pp. 1--6.
\newblock \href {https://doi.org/10.1109/ICNTE.2017.7947882}
  {\path{doi:10.1109/ICNTE.2017.7947882}}.

\bibitem{Piotrowski2020}
A.~P. Piotrowski, J.~J. Napiorkowski, A.~E. Piotrowska, Population size in
  particle swarm optimization, Swarm and Evolutionary Computation 58 (2020)
  100718.
\newblock \href {https://doi.org/https://doi.org/10.1016/j.swevo.2020.100718}
  {\path{doi:https://doi.org/10.1016/j.swevo.2020.100718}}.

\bibitem{Dua2019}
D.~Dua, C.~Graff, \href{http://archive.ics.uci.edu/ml}{{UCI} machine learning
  repository} (2017).
\newline\urlprefix\url{http://archive.ics.uci.edu/ml}

\bibitem{Patricio2018}
M.~Patr{\'\i}cio, J.~Pereira, J.~Cris{\'o}stomo, P.~Matafome, M.~Gomes,
  R.~Sei{\c c}a, F.~Caramelo, Using resistin, glucose, age and bmi to predict
  the presence of breast cancer, BMC Cancer 18~(1) (2018) 29.
\newblock \href {https://doi.org/10.1186/s12885-017-3877-1}
  {\path{doi:10.1186/s12885-017-3877-1}}.

\bibitem{Hoffmann2018}
G.~Hoffmann, A.~Bietenbeck, R.~Lichtinghagen, F.~Klawonn, Using machine
  learning techniques to generate laboratory diagnostic pathways—a case
  study, Journal of Laboratory and Precision Medicine 3~(6) (2018).
\newblock \href {https://doi.org/10.21037/jlpm.2018.06.01}
  {\path{doi:10.21037/jlpm.2018.06.01}}.

\bibitem{Benedek2021}
B.~Rozemberczki, P.~Scherer, O.~Kiss, R.~Sarkar, T.~Ferenci, Chickenpox cases
  in hungary: a benchmark dataset for spatiotemporal signal processing with
  graph neural networks (2021).
\newblock \href {http://arxiv.org/abs/2102.08100} {\path{arXiv:2102.08100}}.

\bibitem{Metzger2015}
A.~Metzger, P.~Leitner, D.~Ivanović, E.~Schmieders, R.~Franklin, M.~Carro,
  S.~Dustdar, K.~Pohl, Comparing and combining predictive business process
  monitoring techniques, IEEE Transactions on Systems, Man, and Cybernetics:
  Systems 45~(2) (2015) 276--290.
\newblock \href {https://doi.org/10.1109/TSMC.2014.2347265}
  {\path{doi:10.1109/TSMC.2014.2347265}}.

\bibitem{DeVito2008}
S.~{De Vito}, E.~Massera, M.~Piga, L.~Martinotto, G.~{Di Francia}, On field
  calibration of an electronic nose for benzene estimation in an urban
  pollution monitoring scenario, Sensors and Actuators B: Chemical 129~(2)
  (2008) 750--757.
\newblock \href {https://doi.org/10.1016/j.snb.2007.09.060}
  {\path{doi:10.1016/j.snb.2007.09.060}}.

\bibitem{Gondek2016}
C.~Gondek, D.~Hafner, O.~R. Sampson, Prediction of failures in the air pressure
  system of scania trucks using a random forest and feature engineering, in:
  Advances in Intelligent Data Analysis XV, Springer International Publishing,
  Cham, 2016, pp. 398--402.

\bibitem{Bouchette2019}
F.~Bouchette, \href{https://oreme.org/observation/ltc/}{{OREME:} the coastline
  observation system} (2019).
\newline\urlprefix\url{https://oreme.org/observation/ltc/}

\bibitem{Snoek2012}
J.~Snoek, H.~Larochelle, R.~P. Adams, Practical bayesian optimization of
  machine learning algorithms, in: F.~Pereira, C.~Burges, L.~Bottou,
  K.~Weinberger (Eds.), Advances in Neural Information Processing Systems,
  Vol.~25, Curran Associates, Inc., 2012.

\bibitem{Carrasco2020}
J.~Carrasco, S.~García, M.~Rueda, S.~Das, F.~Herrera, Recent trends in the use
  of statistical tests for comparing swarm and evolutionary computing
  algorithms: Practical guidelines and a critical review, Swarm and
  Evolutionary Computation 54 (2020) 100665.
\newblock \href {https://doi.org/https://doi.org/10.1016/j.swevo.2020.100665}
  {\path{doi:https://doi.org/10.1016/j.swevo.2020.100665}}.

\bibitem{Laurent2010}
A.~Laurent, B.~Negrevergne, N.~Sicard, A.~Termier, Pgp-mc: Towards a multicore
  parallel approach for mining gradual patterns, in: H.~Kitagawa, Y.~Ishikawa,
  Q.~Li, C.~Watanabe (Eds.), Database Systems for Advanced Applications,
  Springer Berlin Heidelberg, Berlin, Heidelberg, 2010, pp. 78--84.

\end{thebibliography}

\end{document}